\documentclass[10pt,journal,compsoc]{IEEEtran}

\usepackage{multicol}
\usepackage{multirow}
\usepackage{bbm}
\usepackage[dvipsnames, svgnames, x11names, table]{xcolor}
\ifCLASSOPTIONcompsoc
  \usepackage[nocompress]{cite}
\else
  \usepackage{cite}
\fi

\usepackage[pdftex]{graphicx}
\usepackage[normalem]{ulem}
\usepackage{amsmath}
\usepackage{amssymb}
\usepackage{algorithm}
\usepackage{algorithmic}
\usepackage{array}
\usepackage{url}
\usepackage{color}
\usepackage{booktabs}
\usepackage{makecell}
\usepackage{hyperref}
\usepackage{physics}
\usepackage{bm}

\usepackage{gensymb}
\usepackage{mathtools}
\usepackage{gensymb}
\usepackage{xspace}
\makeatletter
\DeclareRobustCommand\onedot{\futurelet\@let@token\@onedot}
\def\@onedot{\ifx\@let@token.\else.\null\fi\xspace}
 
\def\ie{\emph{i.e}\onedot} 
 
 \def\vs{\emph{vs}\onedot}
 
\def\etal{\emph{et al}\onedot}

\ifCLASSOPTIONcompsoc
 \usepackage[caption=false,font=footnotesize,labelfont=sf,textfont=sf]{subfig}
\else
 \usepackage[caption=false,font=footnotesize]{subfig}
\fi

\begin{document}
%
\title{Learning Disentangled Label Representations for Multi-label Classification}
%
%

\author{Jian~Jia,
        Fei~He,
        Naiyu~Gao, 
        Xiaotang~Chen,~\IEEEmembership{Member,~IEEE,}
        and~Kaiqi~Huang,~\IEEEmembership{Senior~Member,~IEEE}


\thanks{J. Jia is with the School of Artificial Intelligence, University of Chinese Academy of Sciences (UCAS), Beijing 100049, China, and also the Center for Research on Intelligent System and Engineering (CRISE), Institute of Automation, Chinese Academy of Sciences (CASIA), Beijing 100190, China (e-mail: jian.jia@outlook.com)}

\thanks{F. He, N. Gao, and X.Chen are with the Center for Research on Intelligent System and Engineering (CRISE), Institute of Automation, Chinese Academy of Sciences (CASIA), Beijing 100190, China (e-mail: hefei2018@ia.ac.cn;gaonaiyu2017@ia.ac.cn;xtchen@nlpr.ia.ac.cn)}

\thanks{K. Huang are with the Center for Research on Intelligent System and Engineering, Institute of Automation, Chinese Academy of Sciences, Beijing 100190, China, and the School of Artificial Intelligence, University of Chinese Academy of Sciences (UCAS), Beijing 100049, China, and the CAS Center for Excellence in Brain Science and Intelligence Technology, Shanghai 200031, China (e-mail: ; kqhuang@nlpr.ia.ac.cn)}

}

%
%

\markboth{Journal of \LaTeX\ Class Files,~Vol.~xx, No.~x, April~2022}%
{Shell \MakeLowercase{\textit{et al.}}: Bare Demo of IEEEtran.cls for Computer Society Journals}
\IEEEtitleabstractindextext{%
\begin{abstract}
Although various methods have been proposed for multi-label classification, most approaches still follow the feature learning mechanism of the single-label (multi-class) classification, namely, learning a shared image feature to classify multiple labels. However, we find this One-shared-Feature-for-Multiple-Labels (OFML) mechanism is not conducive to learning discriminative label features and makes the model non-robustness. For the first time, we mathematically prove that the inferiority of the OFML mechanism is that the optimal learned image feature cannot maintain high similarities with multiple classifiers simultaneously in the context of minimizing cross-entropy loss. To address the limitations of the OFML mechanism, we introduce the One-specific-Feature-for-One-Label (OFOL) mechanism and propose a novel disentangled label feature learning (DLFL) framework to learn a disentangled representation for each label. The specificity of the framework lies in a feature disentangle module, which contains learnable semantic queries and a Semantic Spatial Cross-Attention (SSCA) module. Specifically, learnable semantic queries maintain semantic consistency between different images of the same label. The SSCA module localizes the label-related spatial regions and aggregates located region features into the corresponding label feature to achieve feature disentanglement. We achieve state-of-the-art performance on eight datasets of three tasks, \ie, multi-label classification, pedestrian attribute recognition, and continual multi-label learning.

\end{abstract}


}

\maketitle

\IEEEdisplaynontitleabstractindextext

%
\IEEEpeerreviewmaketitle

\IEEEraisesectionheading{\section{Introduction}\label{sec:introduction}}

%
%
%
%

\IEEEPARstart{A}{s} an expansion of single-label classification \cite{read2011classifier} (\ie, multi-class classification \cite{zhang2013review}) task, the multi-label classification task has attracted increasing attention and achieved remarkable progress due to its wide application in various scenarios, such as image classification \cite{yang2016exploit, liu2021emerging}, audio tag annotation \cite{lo2011cost}, video surveillance \cite{li2018richly}, clinical diagnosis \cite{chen2021multi}, and person re-identification \cite{Wang_2020_CVPR}.
Multi-label classification task aims to predict multiple labels for one image, such as object categories for a natural scene image, human attributes for a pedestrian image, and lung diseases for a chest X-ray image.
Following the framework of single-label classification methods, most multi-label classification methods take the image feature as the label feature. Namely, adopt the average-pooling \cite{ridnik2021asymmetric} or max-pooling operation \cite{chen2019multi, chen2021P_GCN} to extract one shared image feature to classify multiple labels, named the \textbf{\textit{One-shared-Feature-for-Multiple-Label (OFML)}} mechanism in this paper. 

\begin{figure}[ht]
\centering
\includegraphics[width=1.0\linewidth]{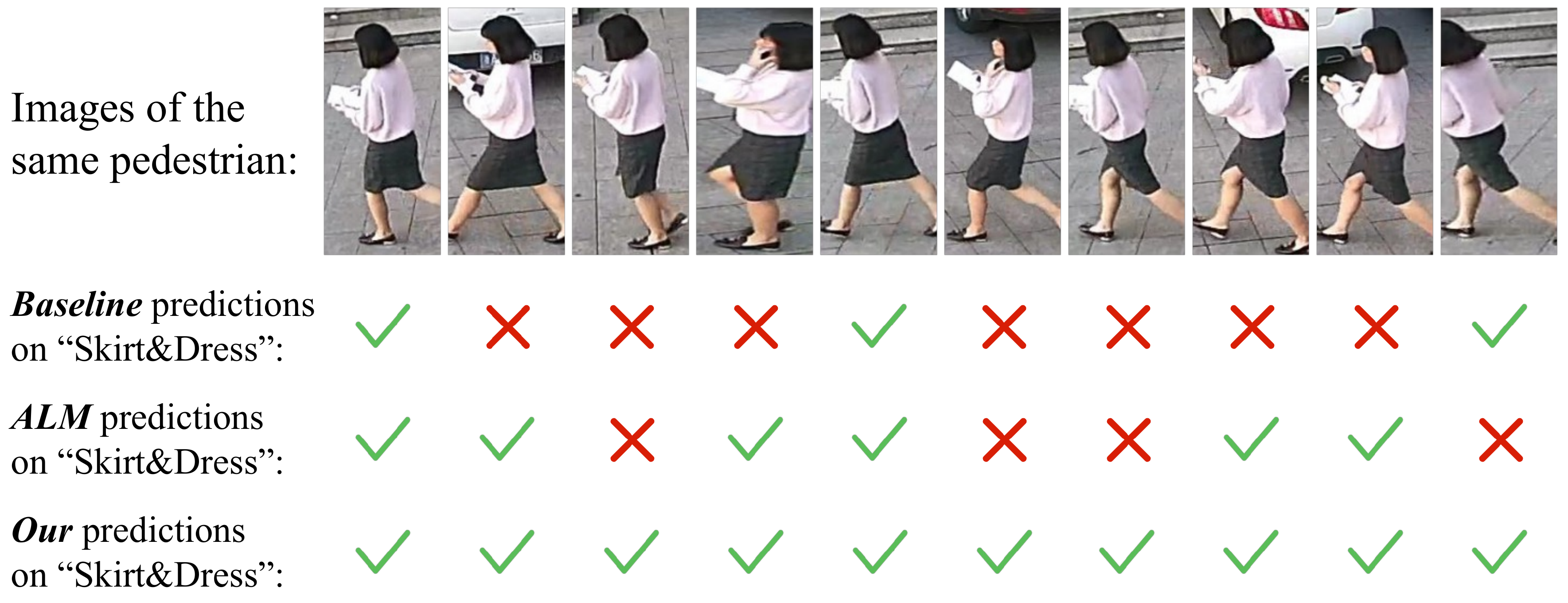}
\caption{\textbf{Illustration of the limitation of the OFML mechanism}. Images of the same pedestrian with the ``Skirt\&Dress" label on the PA-100K \cite{liu2017hydraplus} dataset are listed in the first row. The second and third rows show predictions of baseline methods and the ALM \cite{tang2019Improving}  following the OFML mechanism. Due to the slight variations in background and pedestrian pose, the two methods predict seven and four wrong results for ten images on the ``Skirt\&Dress" label. Following the OFOL mechanism, our method gets the correct predictions for all images.}
\label{fig:inferior_ofml}
\end{figure}

Although plenty of methods based on the OFML mechanism have been proposed and achieved promising performance improvement, there are still some limits. We take the pedestrian attribute recognition task, one of the most prevalent applications of the multi-label classification task in the industry, as an example. As shown in Fig. \ref{fig:inferior_ofml}, for ten images of the same woman wearing a skirt, the baseline method and the ALM \cite{tang2019Improving} \footnote{Predictions of the ALM method \cite{tang2019Improving} is obtained by the code and pre-trained model published in \url{https://github.com/chufengt/iccv19_attribute}.}, both based on the OFML mechanism, give seven and four wrong predictions on the ``Skirt\&Dress" label due to the slight variations in background and pedestrian pose. In contrast, our proposed method gets the correct predictions for all images, regardless of background and pose variations. \textbf{Why does the previous approach produce incorrect judgments because of minor changes in the regions unrelated to the ground truth label?}  
We argue that the OFML mechanism adopted by most previous works \cite{tang2019Improving, sarafianos2018deep, guo2019visual, chen2021P_GCN, chen2019multi, ridnik2021asymmetric, chen2022sst, gao2021MCAR, zhu2021residual} neglects essential characteristics of multi-label classification that distinguish it from the single-label classification task.
Thus, the OFML mechanism is inferior for multi-label classification, and the reasons mainly lie in two aspects.

\begin{figure}[t]
	\centering
    \includegraphics[width=1\linewidth ]{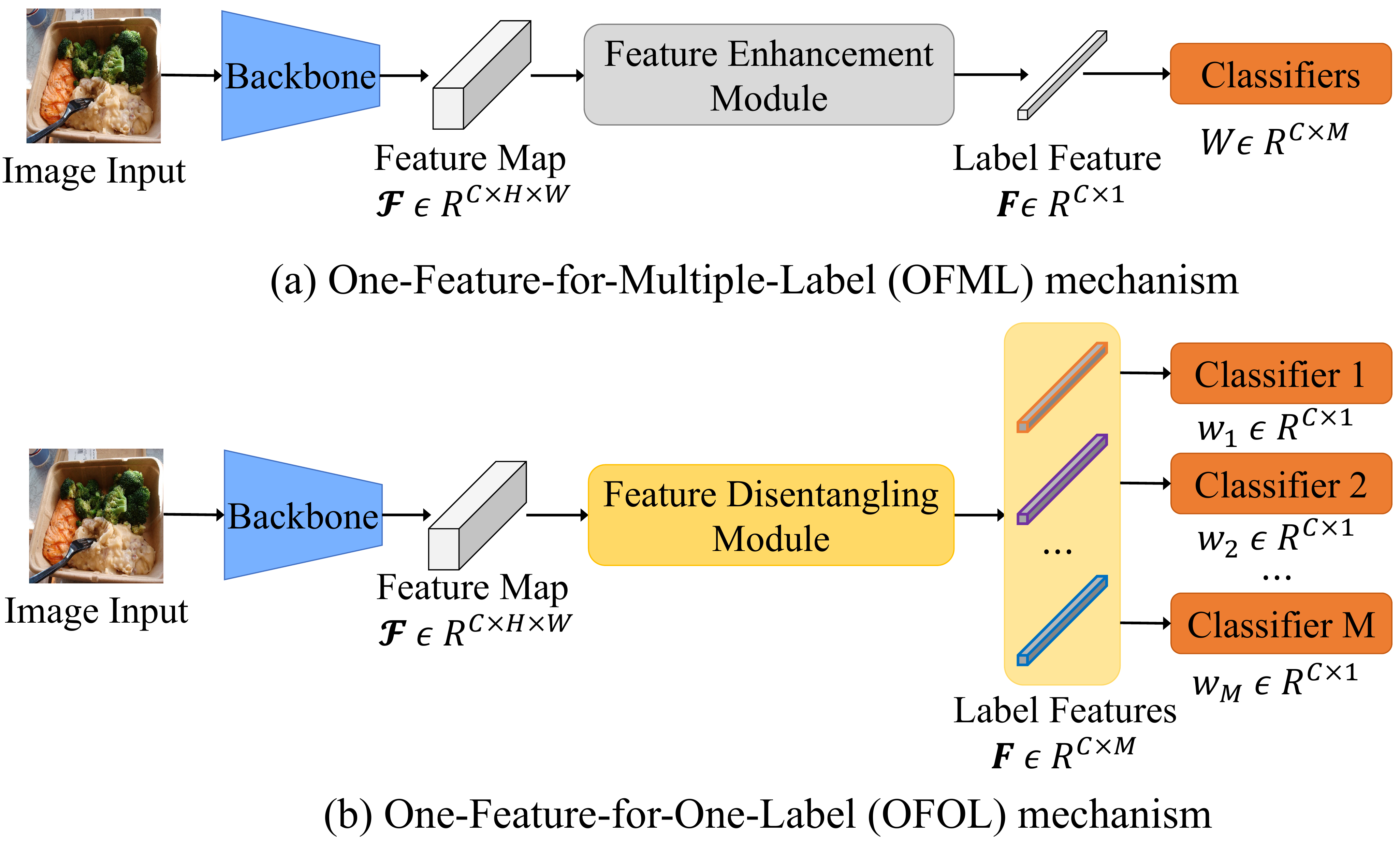}
	\caption{\textbf{Comparison between OFML and OFOL mechanisms}. (a) The OFML mechanism exploits the feature enhancement module to enhance the feature representation and adopts the pooling operation to get the shared label feature for all labels. (b) The OFOL mechanism replaces the feature enhancement module with a feature disentangling module to extract a disentangled feature for each label. }
	\label{fig:OFML_OFOL}
\end{figure}

\begin{figure}[t]
\centering
\includegraphics[width=1.0\linewidth]{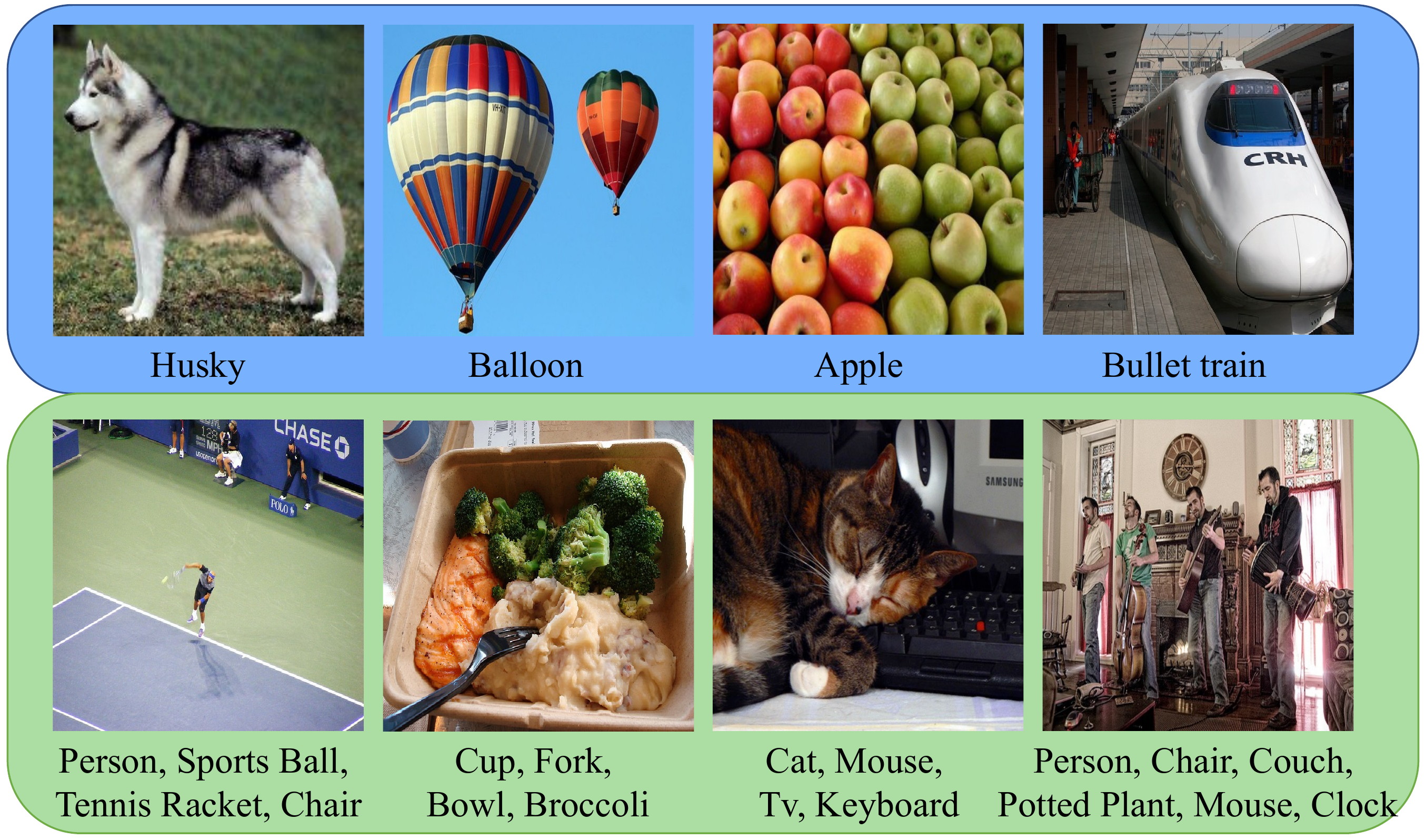}
\caption{\textbf{Task Comparison between single-label and multi-label classification}. Images of Imagenet \cite{deng2009imagenet} and MS-COCO \cite{lin2014microsoft} are examples of single-label and multi-label classification, as listed in the first and second row, respectively. For the single-label classification, one image has only one label (class), and its object occupies the most image space. For the multi-label classification, there are multiple labels (classes) in an image, and the size of the objects of these labels varies. Using the image feature to represent label features for multi-label classification ignores the representation of non-dominant labels, such as ``Sports Ball" in the first image of the second row.}
\label{fig:task_comp}
\end{figure}

First, the pooled image feature cannot effectively represent the feature of each label in the multi-label classification. Following the single-label training pipeline, multi-label approaches \cite{zhu2017learning, tang2019Improving, guo2019visual, chen2021P_GCN, ridnik2021asymmetric, chen2022sst, gao2021MCAR, zhu2021residual} propose various feature enhancement modules to refine the image feature map and adopt the average-pooling or max-pooling operation to obtain the final image feature vector for label classification, as shown in Fig. \ref{fig:OFML_OFOL}(a).
For the single-label classification datasets, such as ImageNet \cite{deng2009imagenet},  one image contains only one dominant category, and objects of that category occupy most of the image space, as shown in Fig. \ref{fig:task_comp}. It is reasonable to take the image feature as the category feature. Therefore, previous methods \cite{simonyan2014very, he2016deep, huang2017densely, hu2018squeeze} proposed for single-label classification adopt the average-pooled image feature as the category feature to conduct classification. 
However, distinguishing from the single-label classification, images of multi-label datasets like COCO \cite{lin2014microsoft} in Fig. \ref{fig:task_comp} contain various objects of multiple labels. 
Although the space occupied by each label in the image varies, labels are equivalent for the multi-label classification task. Image feature obtained by average-pooling (max-pooling) operation like single-label classification inevitably erases the spatial information of labels with small size. It only retains the feature of dominant labels that occupy most of the image space. Thus, it is unreasonable to represent the features of different labels by a shared image feature.

Second, we argue that the image feature learned by the OFML mechanism entangles all label features into one representation. We mathematically prove that this feature learning mechanism hinders the model from learning discriminative features for each label. As we all know, the smaller the angle of the label feature to the corresponding classifier (closer to 0\degree) and the larger the angle to other classifiers (closer to 90\degree), the more discriminative the label feature is. However, we find that, given a fixed feature channel dimension (2048 in ResNet-101 \cite{he2016deep}), angles between the optimal image feature learned by the OFML mechanism and label classifiers converge to 90\degree as the number of labels increases. This phenomenon suggests that the shared image feature learned by the OFML mechanism is not discriminative enough to distinguish all labels. On the other hand, since the decision boundary of each label is perpendicular to the classifier in the multi-label classification, the angle between the feature and the classifier close to 90\degree means that the margin between the feature and the decision boundary is close to 0. This property results that a slight change in the input image can cause the features to cross the decision boundary (perpendicular to the classifier), making the model non-robust, as illustrated in Fig. \ref{fig:inferior_ofml}. The detailed analysis and proof are introduced in Sec.\ref{sec:proof}. As far as we know, our work is the first to mathematically reveal the limitations of the OFML mechanism.

To alleviate the limitations in the OFML mechanism, we introduce the \textbf{\textit{One-specific-Feature-for-One-Label (OFOL)}} mechanism in Fig. \ref{fig:OFML_OFOL}(b) and propose a Disentangled Label Feature Learning (DLFL) framework as illustrated in Fig. \ref{fig:framework}. Instead of using one shared image feature as the feature of all labels, the DLFL framework extracts multiple label-specific features based on precise spatial localization and representative semantic queries. Specifically, we implement the feature disentangling module by learnable semantic queries and a Semantic-Spatial Cross-Attention (SSCA) module. To demonstrate that the superiority of our DLFL framework is brought by the OFML mechanism rather than a complex module and more parameters, we carefully design the SSCA module as straightforward as possible while keeping its powerful feature representation. The learnable semantic queries learn the unique and consistent semantic characteristics of each label from all samples during the training process. Taking an image feature map and semantic queries as inputs, the SSCA module locates label-related spatial regions based on semantic queries for each label. Then, it aggregates the located spatial features into the corresponding label feature. The generated label features can be taken as the semantic queries for the next SSCA module, together with the image feature map to refine label features further. This process can be conducted iteratively and formulates the cascaded SSCA module. The final output feature of each label is sent to the corresponding classifier to get the final prediction.


To verify the effectiveness of the OFOL mechanism and our DLFL framework, we conduct experiments on eight datasets of three tasks and achieve state-of-the-art performance. First, we evaluate our methods on the generic multi-label classification task, which mainly consists of three datasets MS-COCO \cite{lin2014microsoft}, PASCAL-VOC \cite{everingham2010pascal}, and NUSWIDE \cite{chua2009nus}. Second, experiments are conducted on the pedestrian attribute recognition task to validate the superiority of our methods in real-world application scenarios. Pedestrian attribute recognition task replaces the natural scene images and category labels in generic multi-label classification with surveillance video frames and human attributes. PA-100K \cite{liu2017hydraplus}, $\text{PETA}_{\text{ZS}}$ \cite{jia2021rethinking}, and $\text{RAP}_{\text{ZS}}$ \cite{jia2021rethinking} are selected as the pedestrian attribute recognition datasets, since these datasets follow the zero-shot settings between training and test sets and meet the essential requirement of practical industry application. Third, we demonstrate the broad applicability of our approach to the continual multi-label learning datasets $\text{COCO}_{\text{seq}}$ \cite{kim2020imbalanced} and $\text{NUSWIDE}_{\text{seq}}$ \cite{kim2020imbalanced}. As a promising task, continual multi-label learning aims to solve the catastrophic forgetting problem in the sequential multi-label data stream. Significant performance improvement on continual multi-label learning datasets demonstrates that our approach has good extensibility and can be used as a plug-and-play module.

In summary, the main contributions of our work are as follows:

\begin{itemize}
  \item \textbf{We expose the limitations of the OFML mechanism adopted in most existing works}. To minimize the classification loss, the shared image feature learned by the OFML mechanism has to make a trade-off between mutually perpendicular classifiers, resulting in non-discriminative representations and a non-robust model. As far as we know, we are the first to consider the limitations of the OFML mechanism and prove it mathematically.
  \item \textbf{We follow the OFOL mechanism and propose the DLFL framework to solve the issues in the OFML mechanism}. The DLFL framework consists of the SSCA module and semantic queries to locate label-related spatial regions and extract a specific feature for each label.
  \item \textbf{We conduct extensive experiments on eight datasets of three tasks to validate the superior of our method}. We evaluate our method on generic multi-label classification, pedestrian attribute recognition, and continual multi-label learning tasks and achieve state-of-the-art performance on all tasks.
\end{itemize}

Compared to the preliminary version presented as a conference paper \cite{jia2022learning}, we have made substantial improvements in technical contributions and exhaustive experiments and updated the state-of-the-art performance.
First, we have further elaborated the module based on the published version to achieve higher performance with fewer parameters.
Second, in addition to experiments on pedestrian attribute dataset PA-100K \cite{liu2017hydraplus} reported in the conference paper, we conduct comprehensive experiments on other seven datasets of three tasks to demonstrate the broad applicability of our method. 
Third, we elaborate on the reasons for the effectiveness of our method and give concrete mathematical proof. Lastly, we achieve higher performance and update the state-of-the-art performance compared to our conference version \cite{jia2022learning}.

The rest of this paper is organized as follows. Section \ref{sec:related_work} provides a review of works in three related tasks. Section \ref{sec:proposed_method} introduces the prerequisite knowledge of the multi-label classification task, describes the details and limitations of the OFML mechanism, and proposes our DLFL framework. Exhaustive studies and experiments of three tasks are shown in Section \ref{sec:experiments} to validate the effectiveness of our method. Finally, we conclude this paper and discuss the future application of our paper in Section \ref{sec:conclusion}.








\section{Related Work} \label{sec:related_work}
In this section, we review the previous works from three perspectives. First are the works related to the generic multi-label classification, the most common task for evaluating multi-label methods. Second are the studies about pedestrian attribute recognition, which applies multi-label classification in video surveillance scenarios. Finally, progress on continual multi-label learning, an emerging task, is introduced.

\subsection{Generic multi-label classification}

Generic multi-label classification methods can be divided into two categories based on their methodology. One category of methods focuses on learning label-related spatial features, and the other aims to construct label correlation.

In learning label-related spatial features, Wei \etal \cite{wei2015hcp} extracted an arbitrary number of object bounding boxes from the input image as the hypotheses and aggregated hypotheses prediction by max-pooling to get the final results. Unlike using a pre-trained detector to produce the object proposal \cite{wei2015hcp}, Wang \etal \cite{wang2016beyond} exploited the stochastic scaling and cropping to generate input augmentation. To enhance the spatial localization ability of the model, Zhu \etal \cite{zhu2017learning} designed the spatial regularization network to learn spatial attention maps for each label. Similarly, Wang \etal \cite{wang2017multi} introduced the Spatial Transformer module \cite{jaderberg2015spatial} to learn spatial attention regions for each label by four parameters.  
Gao \etal \cite{gao2021MCAR} designed a two-stream network, where the global stream located object regions and the local stream extracted object features.
Zhu \etal \cite{zhu2021residual} proposed the residual attention to dynamically combine the features generated by average pooling and max pooling operations.
Zamir \etal \cite{ridnik2021asymmetric} designed an asymmetric loss to solve the positive-negative imbalance problem in multi-label classification. 
Inspired by the success of the Transformer \cite{vaswani2017attention}, Chen \etal \cite{chen2022sst} proposed a spatial transformer and semantic transformer to learn powerful feature representations by simultaneously capturing spatial and semantic correlations.

In label dependency construction, Wang \etal \cite{wang2017multi} utilized LSTM \cite{hochreiter1997long} to capture the global context between label features and sequentially generated predictions based on discriminative label regions output by the Spatial Transformer module \cite{jaderberg2015spatial}.  
Chen \etal \cite{chen2019learning} decoupled semantic-related features and adopted a gated recurrent update mechanism to propagate messages based on the Graph Neural Network (GNN) \cite{li2015gated}.
Chen \etal \cite{chen2019multi, chen2021P_GCN} introduced the Graph Convolution Network (GCN) \cite{kipf2017semi} to capture the label correlation and generate the classifier based on the label correlation matrix.
Ye \etal \cite{ye2020attention} proposed a dynamic GCN module to capture content-aware category relations for each image.
Zhao \etal \cite{zhao2021transformer} proposed a structural and semantic relation graph to capture long-range correlations and model semantic characteristics.
Lanchantin \etal \cite{lanchantin2021general} introduced the Transformer into multi-label classification and designed a label mask training pipeline to learn label correlations.

Our work falls into the first category and focuses on learning label-related spatial features. Previous approaches either intuitively adopt the OFOL mechanism \cite{zhu2017learning, wang2017multi, chen2019learning, gao2021MCAR, zhao2021transformer,chen2022sst} without giving a reason or follow the inferior OFML mechanism \cite{chen2021P_GCN, zhu2021residual, ridnik2021asymmetric}. In contrast to previous work, we provide an in-depth analysis of the OFML mechanism and reveal its limitations, as well as elaborate and justify why the OFOL mechanism works. Besides, compared with other methods following the OFOL mechanism, our method achieves comparable, even better performance with fewer parameters and computations.

\subsection{Pedestrian attribute recognition}

Since Li \etal \cite{li2015deepmar} introduced deep learning into pedestrian attribute recognition, various methods have been proposed, and significant progress has been made. We divide current works into two categories according to the feature used to classify attributes.

The first category of methods \cite{li2015deepmar, yu2016weakly, liu2017hydraplus, li2018pose, liu2018localization, sarafianos2018deep, han2019attribute, guo2019visual, tan2020relation} extracted a shared global feature to classify all attributes. These methods design various feature enhancement modules by adopting attention mechanisms and taking the average-pooled image feature as the global feature. Liu \etal \cite{liu2017hydraplus} proposed a multi-direction attention network, HydraPlus-Net, to utilize diverse semantic attention from different layers. 
Sarafianos \etal \cite{sarafianos2018deep} constructed a Visual Attention and Aggregation (VAA) module and applied it on multi-scale feature maps. 
Guo \etal \cite{guo2019visual} proposed attention consistency loss to align the attention regions of augmentations of the same image. 
Tan \etal \cite{tan2020relation} proposed a two-branch network JLAC, where the ARM branch generated attribute features based on the average-pooled image feature and the CRM branch concatenated graph node features to make predictions. 
Jia \etal \cite{jia2021spatial} utilized the human prior to impose the spatial and semantic regularization on the feature learning process.

The second category of methods learned multiple features to classify the attributes. 
Zhao \etal \cite{zhao2018grouping} introduced human key points to generate body proposals and adopted RoI average pooling layers to extract proposal features as the group features. 
Instead of utilizing human key points, Li \etal \cite{li2019pedestrian} exploited the human parsing model to locate body regions and adopted graph convolution networks to obtain corresponding group features. 
Lin \etal \cite{wang2017attribute} took horizontally divided features as input and introduced LSTM \cite{hochreiter1997long} to decode individual features for each attribute. 
Li \etal \cite{li2019visual} proposed a visual-semantic graph reasoning framework to capture spatial region relations and attribute semantic relations. Node features of the graph networks are used as the specific features for each attribute. 
Tang \etal \cite{tang2019Improving} constructed multi-scale feature maps by Feature Pyramid Network (FPN) \cite{lin2017feature} and proposed the attribute location module (ALM) to extract features for each attribute in each scale feature map.

Compared to the above methods, our work can learn a representative feature for each attribute without human prior \cite{wang2017attribute}, such as key points \cite{zhao2018grouping} and human parsing \cite{li2019pedestrian}. As an industry application of multi-label classification in intelligent surveillance, our analysis of the limitations in the OFML mechanism also applies to pedestrian attribute recognition, which is not mentioned in previous works.

\subsection{Continual learning}

As a promising approach, continual learning addresses catastrophic forgetting by preserving learned knowledge of previous tasks during training on the new tasks \cite{li2017learning}. Many works investigate continual learning in different computer vision tasks. For the semantic segmentation task, Yang \etal \cite{yang2022uncertainty} integrated contrastive learning with a knowledge distillation framework to solve catastrophic forgetting. For the few-shot classification task, Zhao \etal \cite{zhao2021mgsvf} proposed disentangling frequency space to achieve a good balance between the slow forgetting of old knowledge and fast adaptation to new knowledge. For the objection detection task, Zhao \etal \cite{kj2021incremental} designed a gradient-based meta-learning approach to minimize information forgetting and maximize knowledge transfer across incremental tasks. For video action recognition,
Park \etal \cite{park2021class} exploited the temporal information and proposed an importance mask to preserve the frame with transferrable knowledge for video action recognition. 
Based on the replay-based approach,  Kim \etal \cite{kim2020imbalanced} proposed continual multi-label learning settings and designed a sampling strategy to maintain more samples of minority classes to alleviate the long-tailed distribution problem. Our work follows the continual multi-label learning settings proposed by \cite{kim2020imbalanced} to explore the feasibility of our solution in the continual learning setting.

\section{Proposed Method} \label{sec:proposed_method}

\begin{figure*}[th]
	\centering
    \includegraphics[width=0.9\linewidth ]{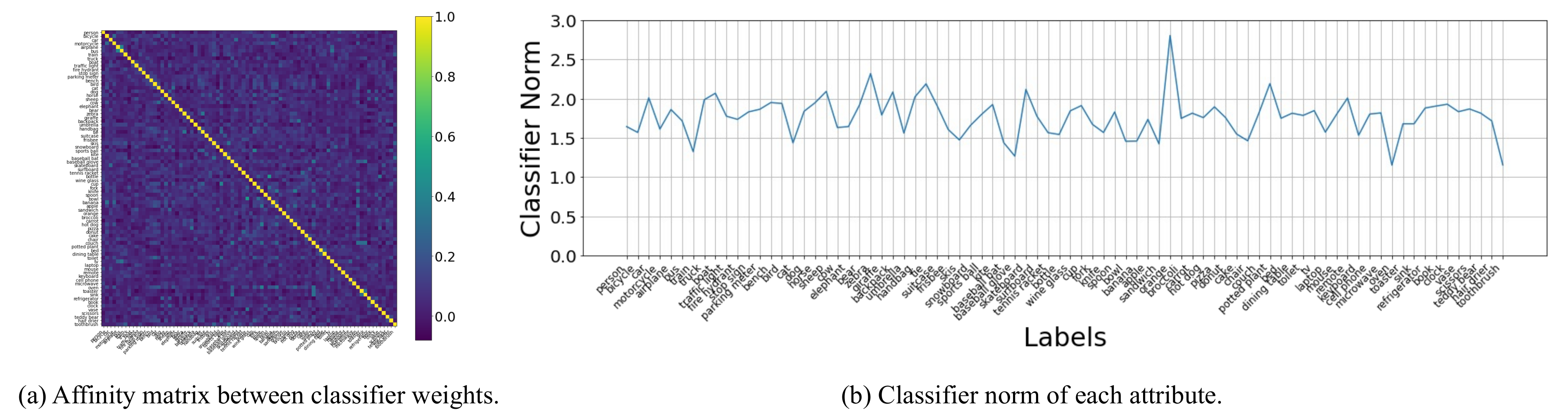}
	\caption{{\bf Illustration of two experimental observations}. In (a), we show the affinity matrix between normalized classifier weights. In (b), we count the classifier norm of each label. The axes of the affinity matrix and x-axis in (b) list the category names (Best viewed in magnification). We can conclude that classifier weights are mostly orthogonal, and the norms of classifier weights fluctuate in a small range. }
	\label{fig:observation}
	\vspace{-1em}
\end{figure*}

In this section, we first introduce the prerequisite knowledge of the multi-label classification task. According to the common settings and the inference process, we point out that the only determinant for label prediction is the angle between the label feature vector and the corresponding classifier weight. Then, we mathematically expose the limitations of the One-shared-Feature-for-Multiple-Label (OFML) mechanism adopted by most methods and introduce the One-specific-Feature-for-One-Label (OFOL) mechanism to tackle the deficiencies in OFML. Finally, following the OFOL mechanism, we construct the Disentangled Label Feature Learning (DLFL) framework and design a Semantic Spatial Cross-Attention (SSCA) module to achieve feature disentanglement.

\subsection{Basics of multi-label classification}

Given a dataset $\mathbb{D} = \{x_{i}, y_{i}, | i=1, \dots, N \}$ and $y_{i} \in \{0, 1\}^{M}$, a multi-label classification model aims to learn discriminative label features and predict multiple labels $\hat{y}_{i}$ for an image $x_{i}$. 
In the inference stage, prediction results of multi-label images depend on the choice of a probability threshold $p_{t}$.
To make an intuitive prediction and a fair comparison, all existing methods set the probability threshold $p_{t} = 0.5$. Therefore, for the $j$-th label of the $i$-th image $x_{i}$, the prediction result $\hat{y}_{i, j}$ is decided as follows:
\begin{align} \label{eq:yij}
	\hat{y}_{i, j} & = 
    \begin{cases}
      1, & p_{i, j} >= p_{t} \\
      0, & p_{i, j} < p_{t} 
    \end{cases},  \\
     p_{i, j} & = \sigma(logits_{i, j}) ,  \label{eq:pij}
\end{align}
where $p_{i, j}$ is the predicted probability, and $\sigma(\cdot)$ is the sigmoid function $\sigma(x) = \frac{1}{ 1+ e^{-x}}$ . The output of the classifier layer is denoted as $logits$, which is computed as:
\begin{equation} \label{eq:logitsij}
  logits_{i, j} = w^{T}_{j}f_{i} = \|w_{j}\|  \cdot \|f_{i, j}\| \cdot \cos{\theta},
\end{equation}
where $f_{i, j} \in \mathbb{R}^{C} $ is the $j$-th label feature of the sample $x_{i}$, and $W = \{w_{j}|j=1, \dots, M \} \in \mathbb{R}^{C \times M} $ is the classifier weight matrix. Taking Eq. \ref{eq:pij} and Eq. \ref{eq:logitsij} into Eq. \ref{eq:yij}, we conclude that the prediction of the $j$-th label only depends on the sign of $\cos \theta$ in Eq. \ref{eq:logitsij}, \ie, the angle $\theta$ between the label feature $f$ and the classifier weight $w$:
\begin{align} \label{eq:angle}
	\hat{y}_{i, j} & = 
    \begin{cases}
      1, & 0\degree  <= \theta <= 90 \degree \\
      0, & 90\degree < \theta < 180\degree 
    \end{cases} .
\end{align}

Therefore, for a target label, a well-trained model should make angles between positive label features and the corresponding classifier weight as small as possible, or even close to 0\degree, which means high-confidence prediction. Conversely, angles between negative label features and the corresponding classifier weight should be as large as possible, tending to be 180\degree. We use positive (negative) label features to indicate the label features learned from images with (without) corresponding labels. It is worth noting that, for the OFML mechanism, a shared average-pooled or max-pooled image feature $f_{i}$ is taken as the label features for all labels, namely, all label features are the same. Thus, we adopt the $f_{i}$ to denote the label feature learned by the OFML mechanism. For the OFOL mechanism, a specific feature is learned for each label and denoted by $f_{i, j}$ for the $j$-th label feature of the $i$-th image.

\subsection{Review the OFML Mechanism} \label{sec:limit}

\subsubsection{Two experimental observations.} \label{sec:two_observ}

We have two critical experimental observations for a well-trained model that follows the OFML mechanism.

Specifically, one is that most classifier weights of labels are orthogonal. We normalize the classifier weights and compute the affinity matrix between the normalized classifier weights.  As shown in Fig. \ref{fig:observation}(a), almost all elements of affinity matrix are distributed around 0, except for diagonal elements and affinities between semantically opposite labels, such as ``ShortSleeve" with ``LongSleeve" and ``AgeOver60" with ``Age18-60", ``AgeLess18". This phenomenon is also verified in other literature \cite{vershynin2018high}. The other is that classifier norms fluctuate within a small range and are approximately the same, as shown in Fig. \ref{fig:observation}(b). Considering the orthogonality between label classifiers, it is intuitively impossible to make angles between one shared image feature learned by the OFML mechanism and multiple label classifier weights all close to 0\degree. However, these angles are expected to be as small as possible, which implies high-confidence predictions of the model. What is the theoretical optimal angle?

\begin{figure*}[h]
	\centering
	\subfloat[Two-label classification.]{
		\includegraphics[width=0.26\linewidth ]{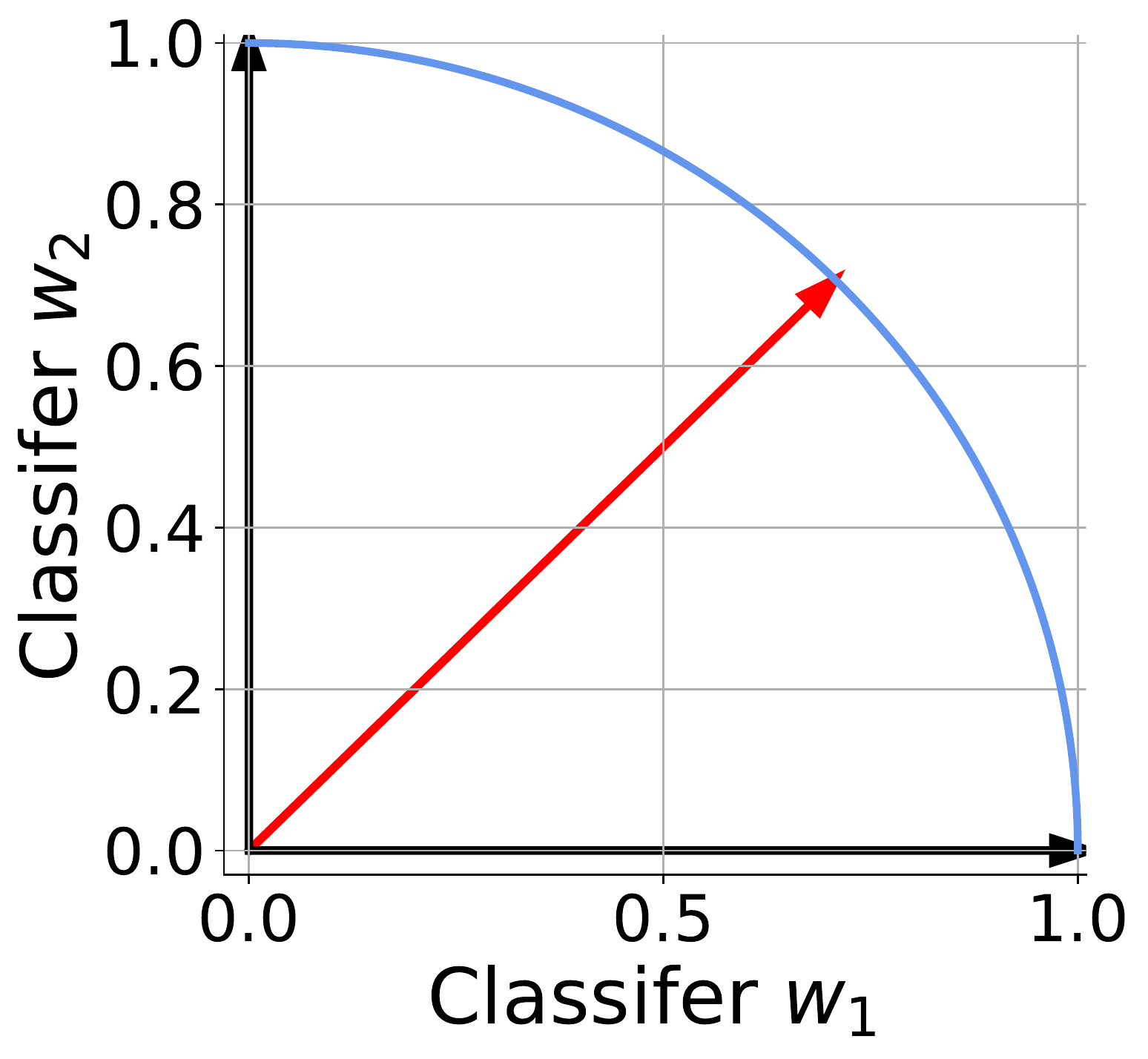} \hspace{0.01\linewidth}} 
	\subfloat[Three-label classification]{
		\includegraphics[width=0.28\linewidth ]{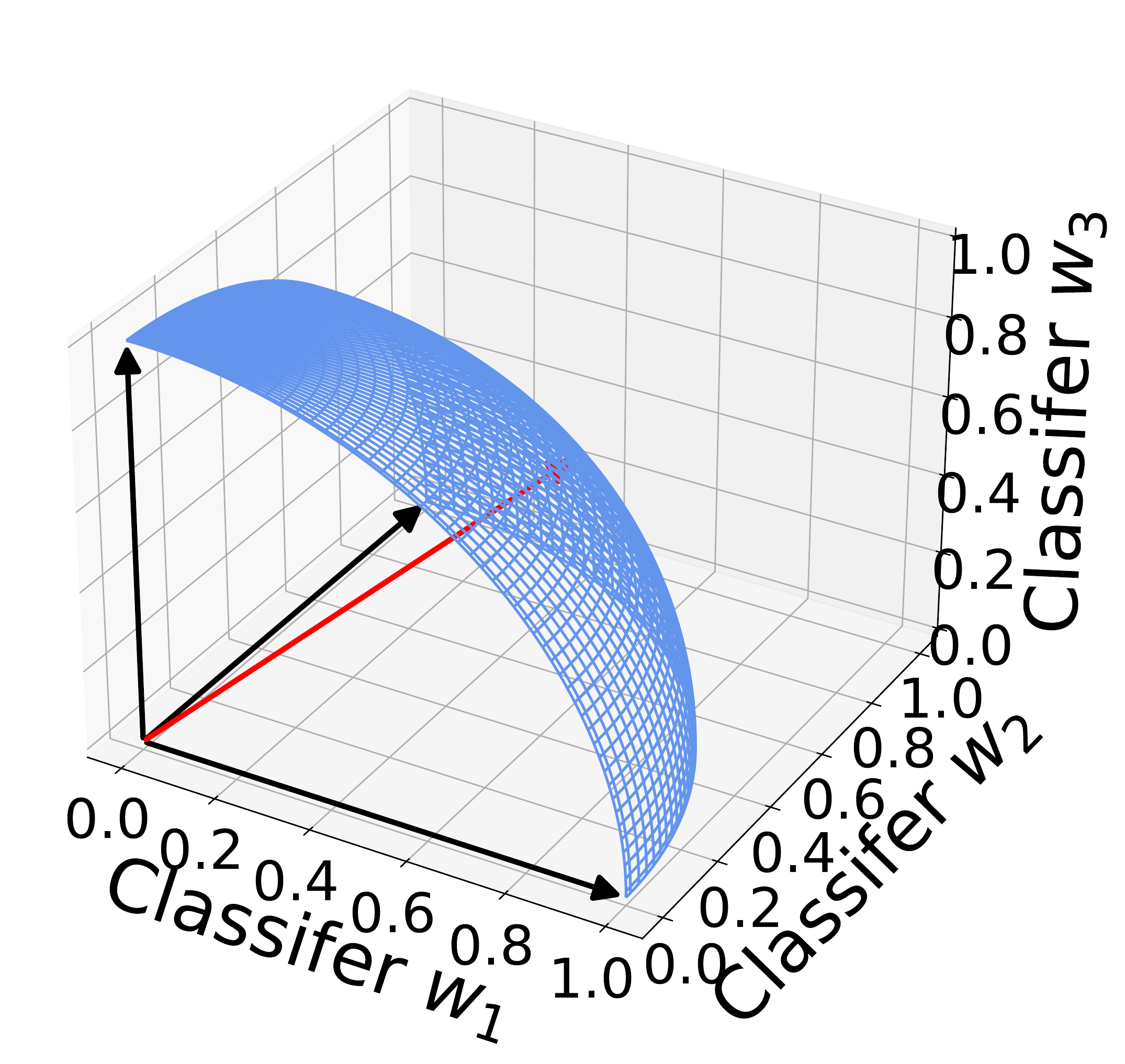} \hspace{0.01\linewidth}}
	\subfloat[Multi-label classification]{
		\includegraphics[width=0.38\linewidth ]{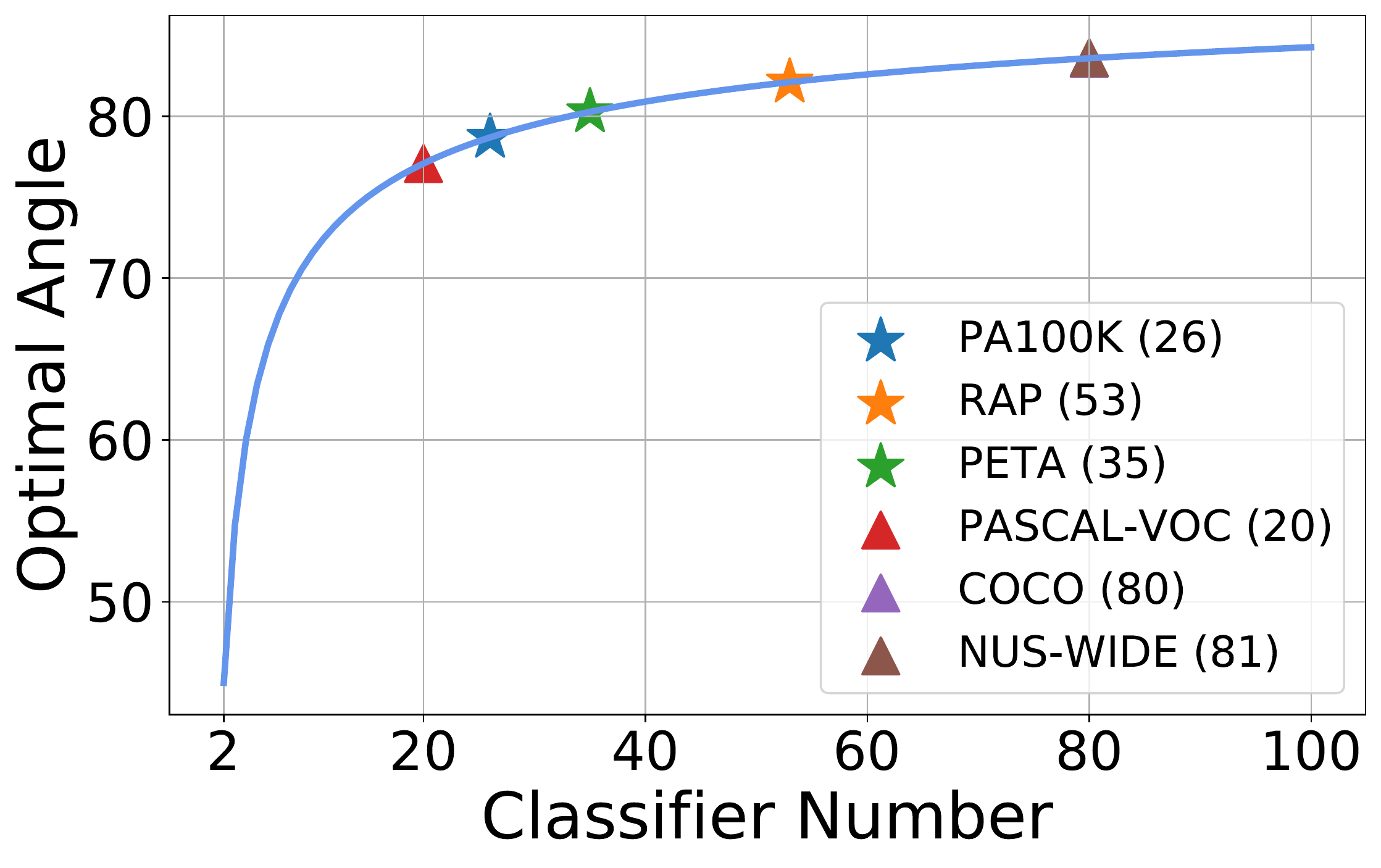}}
	\caption{\textbf{The optimal angle between a shared feature vector and multiple classifier weights}. Assuming that the classifier weights are orthogonal and the classifier norms are identical, the optimal angle is the angle that minimizes the binary cross-entropy loss. We normalize the feature vector (red arrow) and label classifier weights (black arrows) for simplification. In (a) and (b), we depict the optimal angles between a shared feature vector and two classifiers as well as three classifiers, respectively. In (c), we plot the curve of the optimal angle as the number of classifiers increases and give the optimal angles on six popular datasets. Due to the number of labels in COCO \cite{lin2014microsoft} and NUS-WIDE \cite{chua2009nus} are very close (80 \vs 81), the two triangles representing the optimal angle in (c) overlap almost exactly.}
	\label{fig:angle}
\end{figure*}

\subsubsection{Proof of the optimal feature in the OFML mechanism} \label{sec:proof}


Given a sample $(f_{i}, y_{i})$ where $f_{i}$ is the image feature of the $i$-th sample and $y_{i} \in \{0, 1\}^{M}$ is the ground truth label, we assume the number of positive labels (value 1 in ground truth) and the number of negative labels (value 0 in ground truth) are $M_{p}$ and $M_{n}$ where $M = M_{p} + M_{n}$. This means that $M_{p}$ labels are annotated as appearing in the $i$-th image, while the other $M_{n}$ labels do not appear in that image. Following the common practice, binary cross-entropy loss is adopted as the classification loss and computed as:
\begin{align}
    L_{i} &= - \sum_{j=1}^{M} (y_{i,j} \log(p_{i,j})  + (1 - y_{i,j})\log(1 - p_{i,j})) \\ 
    &= - \sum_{j=1}^{M_{p}}\log(p_{i,j}) - \sum_{j=M_{p} + 1}^{M_{p} + M_{n}} \log(1 - p_{i,j}) \\
    &= - \sum_{j=1}^{M_{p}}\log(\sigma(w^{T}_{j}f_{i})) - \sum_{j=M_{p} + 1}^{M_{p} + M_{n}} \log(1 - \sigma(w^{T}_{j}f_{i})) \\ 
    &= - \log \prod_{j=1}^{M_{p}}\sigma(w^{T}_{j}f_{i}) \prod_{j=M_{p} + 1}^{M_{p} + M_{n}}(1 - \sigma(w^{T}_{j}f_{i})). \label{eq:log_sigmoid}
\end{align}

Taking the sigmoid function into Eq. \ref{eq:log_sigmoid}, the loss function is transformed to:
\begin{align}
    L_{i} &= - \log \frac{1}{\prod_{j=1}^{M_{p}} (1 + e^{-w^{T}_{j}f_{i}})  \prod_{j=M_{p} + 1}^{M_{p} + M_{n}}  (1 + e^{w^{T}_{j}f_{i}}) }. 
\end{align}
To minimize the binary cross-entropy loss, we formulate the optimization problem as follows:
\begin{align} \label{eq:theta}
    & \min_{f_{i}} \prod_{j=1}^{M_{p}} (1 + e^{-w^{T}_{j}f_{i}})  \prod_{j=M_{p} + 1}^{M_{p} + M_{n}}  (1 + e^{w^{T}_{j}f_{i}})
\end{align}

According to the second experimental observations  $ \| w_{i} \|_{2} \approx \| w_{j} \|_{2}$  (\textbf{\textit{the classifier norms are almost the same}}), we hypothesis that the production $\alpha$ between image feature norm $\|f_{i}\|_{2}$ and classifier weight norms $\|w_{j}\|_{2}$ are same for the $i$-th sample and adopt a constant $\alpha$ to denote the production result, \ie, $\alpha = \| w_{j} \|_{2} \cdot \|f_{i}\|_{2}$, where $j \in {1,2, \dots, M}$. Therefore, Eq. \ref{eq:theta} is changed to:
\begin{align}
    & \min_{\theta_{i,j}} \prod_{j=1}^{M_{p}} (1 + e^{- \alpha \cos{\theta_{i,j}}})  \prod_{j=M_{p} + 1}^{M_{p} + M_{n}}  (1 + e^{ \alpha \cos{\theta_{i,j}}}) \label{eq:optim_theta}
\end{align}
where $\cos \theta_{i,j} = \frac{w_{j}^{T} \cdot f_{i}}{\| w_{j} \|_{2} \cdot \|f_{i}\|_{2}}$ indicates the angle between the image feature $f_{i}$ and classier weight $w_{j}$.

Taking the image feature dimension $C=2048$ of backbone network ResNet-101 \cite{he2016deep} as an example, we assume the unit vector of feature $f_{i}$ as $\frac{f_{i}}{ \| f_{i} \|_{2} } = \{x_{1}, x_{2}, \dots, x_{2048}\}$. According to the first experimental observations (\textbf{\textit{the classifier norms are almost orthogonal}}), the unit vector of classifier weights can be denoted as $\frac{w_{j}}{ \| w_{j} \|_{2} } = \{0, \dots, 1, \dots, 0\} \in R^{C}$ for simplification where $j$-th value of $\frac{w_{j}}{ \| w_{j} \|_{2}}$ is 1. The cosine function can be concluded as $\cos \theta_{i,j} = \frac{w_{j}}{\| w_{j} \|_{2}} \cdot \frac{f_{i}}{\|f_{i}\|_{2}} = x_{j} $. Thus, the optimization problem \ref{eq:optim_theta} can be transformed into:
\begin{align} \label{eq:optim_xj}
   \min_{x_{j}} & \prod_{j=1}^{M_{p}} (1 + e^{- \alpha x_{j}})  \prod_{j=M_{p} + 1}^{M_{p} + M_{n}}  (1 + e^{ \alpha x_{j}}) \\ 
    s.t. & \sum_{j=1}^{M} x_{j}^{2} = 1, \: \text{and} \:
   \begin{cases}
      0 < x_{j} \leq 1, \: \: 1 \leq j \leq M_{p} \nonumber \\
      -1 \leq x_{j} < 0, \: M_{p} < j \leq M  \nonumber
   \end{cases}       
\end{align}
According to the Lagrange Multiplier and the Chain Rule, Eq. \ref{eq:optim_xj} get the minimum value when $x_{j}$ satisfies the following conditions:
\begin{align} \label{eq:optim_result}
   x_{j} = \begin{cases}
      \frac{\sqrt{M}}{M}, & j = 1, 2, \dots, M_{p} \nonumber \\
      - \frac{\sqrt{M}}{M}, & j =  M_{p}+1,  M_{p}+2, \dots, M_{p}+M_{n} \nonumber
   \end{cases}       
\end{align}
Thus, the optimal shared feature is $\frac{f_{i}}{ \| f_{i} \| } = \{ \frac{\sqrt{M}}{M} , \frac{\sqrt{M}}{M}, \dots, - \frac{\sqrt{M}}{M}, - \frac{\sqrt{M}}{M}, 0, \dots, 0 \}$. 

We take two-label classification task as an example to elaborate our proof. Given a sample $(f_{i}, y_{i})$ where $f_{i}$ is the image feature of the $i$-th sample and $y_{i} \in \{0, 1\}^{2}$ is the ground truth label. We assume that both two labels exists in the $i$-th image, \ie, $y_{i} = \{ 1, 1 \}$, and the classification on other cases can be analyzed in the same way. According to the Eq. \ref{eq:optim_result}, the number of positive labels is $M_{p} = 2$ and the unit vector of the optimal feature is $\frac{f_{i}}{ \| f_{i} \| } = \{ \frac{\sqrt{2}}{2}, \frac{\sqrt{2}}{2}, 0, \dots, 0 \}$. The two orthogonal classifier weights can be denoted as  $w_{1} = \{1, 0, \dots, 0\}$ and $w_{2} = \{0, 1, \dots, 0\}$. As a result, the optimal $\theta_{1} = \theta_{2} = 45\degree$, \ie, \textbf{\textit{the shared feature $f_{i}$ has the same distance from both two classifiers $w_{1}$ and $w_{2}$ }}. We visualize this result in Fig. \ref{fig:angle}(a), where the optimal feature $f_{i}$ is located in the middle of the two classifiers.

\begin{figure*}[ht]
\centering
\includegraphics[width=0.9\linewidth]{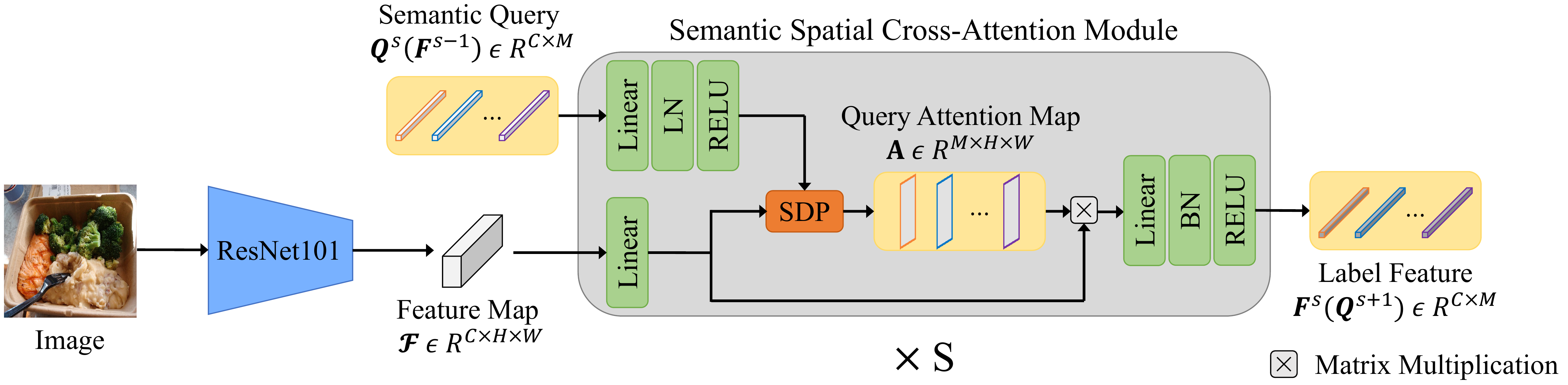}
\caption{\textbf{Illustration of the Disentangled Label Feature Learning (DLFL) framework}. The proposed framework mainly consists of learnable semantic queries $\bm{Q}$ and Semantic Spatial Cross-Attention (SSCA) module. Taking the image feature map $\mathcal{F}$ and semantic query $\bm{Q}^{s}$ as input, the SSCA module locates label-related regions for each label and generates the query attention map $\bm{A}$. Based on the query attention map, located region features are aggregated into label features $\bm{F}^{s}$. The label features $\bm{F}^{s}$ of the current SSCA module can be taken as the semantic query $\bm{Q}^{s+1}$ of the next SSCA module to further refine the label features. }
\label{fig:framework}
\end{figure*}

For the three-label classification task where $y_{i} = \{ 1, 1, 1 \}$, the optimal feature has the same distance from all three classifiers, \ie, the optimal angles are 54.74\degree, as shown in Fig. \ref{fig:angle}(b). Furthermore, for generic multi-label classification on $M$ labels, we conclude that the optimal feature achieves a trade-off in the distance between itself and multiple classifiers to minimize the binary cross-entropy loss. This property makes the optimal angle converge to 90\degree as the number of labels $M$ increases. For example, on generic multi-label datasets MS-COCO \cite{lin2014microsoft} and pedestrian label datasets PA-100K \cite{liu2017hydraplus}, the number of labels $M$ are 80 and 26. According to the Eq. \ref{eq:optim_result}, the optimal angles are 83.58\degree and 78.69\degree respectively, as shown in Fig. \ref{fig:angle}(c). This theoretical conclusion is verified in the experimental statistics in Fig. \ref{fig:angle_dist}. 
.


\subsubsection{Limitations of the OFML Mechanism}


However, the optimal angles close to 90\degree in the training stage are far from our expectations. As a result, a small perturbation can cause features of the test set to cross the decision boundary, causing angles of the test set to be greater than 90\degree and yielding wrong predictions. For example, in the pedestrian attribute recognition task, the learned features are susceptible to changes in pedestrian pose, illumination, and background, resulting in incorrect classifications. Some examples are given in Fig. \ref{fig:inferior_ofml}. Thus, we conclude that the shared features learned by the OFML mechanism are too far from the classifiers, which reduces the robustness of the model on the test set. In addition, it is worth noting that this deficiency is determined by the OFML mechanism and independent of the specific method. 

Besides to large angles between the shared feature and classifiers, there is another limitation of the OFML mechanism. The shared global features, used to classify multiple labels, are average pooled from the image feature map. Nevertheless, the pooling operation erases the spatial distinction of labels and only exploits the channel information. Therefore, the OFML mechanism is inadequate to exploit the differences between labels to learn discriminative features for each label.

\subsection{Proposed DLFL Framework}

To solve the limitations in the OFML mechanism, we propose a Disentangled Label Feature Learning (DLFL) framework following the OFOL mechanism as illustrated in Fig. \ref{fig:framework}. Instead of using a feature enhancement module to extract a shared image feature as the label feature, the DLFL framework relies on a feature disentangling module to disentangle label features from one image feature map. Considering the different spatial distribution of labels in the image, we propose to locate label-related spatial regions and aggregate discriminative regional features as a specific feature for each label. To precisely locate label-related spatial regions, we design learnable semantic queries to guide the spatial localization for each label. The learnable semantic queries exploit the semantic consistency of labels between different samples. For generic multi-label classification, whether the picture is of a man, a woman, a child, or the elderly, intrinsic semantic features of the ``person'' label should be consistent. For pedestrian attribute recognition, whether the pedestrian is wearing a beret, helmet, bucket hat, or baseball cap, intrinsic semantic features of the ``hat" attribute should not be changed.




Therefore, we implement the feature disentangling module with a Semantic Spatial Cross-Attention (SSCA) module. The SSCA module first locates precise spatial regions for each label according to the learnable semantic queries, then aggregates features of located regions as the label features. Concretely, the SSCA module takes semantic query $\bm{Q} \in R^{C \times M}$ and image feature map $\mathcal{F} \in R^{C \times H \times W }$ as inputs, where $M$ is the number of labels. The channel, height, and width dimension of the feature map is represented by $C$, $H$, and $W$ respectively. The image feature map $\mathcal{F}$ is output by the backbone network like ResNet-101 \cite{he2016deep}. Then, query attention maps $\bm{A}$ are generated to locate the spatial regions of each label as follows:
\begin{align}
    \bm{A} = Softmax(\frac{\theta(\bm{Q})^{T}\phi(\mathcal{F})}{\sqrt{C}}),
\end{align}
where $\theta(\bm{Q}) = \bm{W}_{\theta}\bm{Q}$ and $\phi(\mathcal{F}) = \bm{W}_{\phi}\mathcal{F}$ are two linear embedding functions. Query attention map highlights label-related spatial regions and can be used as the affinity matrix to aggregate region features as specific label feature $F^{s}$, which is formulated as:
\begin{align}
    \bm{F}^{s} = \bm{A}\phi(\mathcal{F})^{T},
\end{align}
To further locate reliable spatial regions and refine the label feature, we cascade multiple SSCA modules and take label feature $\bm{F}^{s-1}$ of the previous SSCA module as the semantic query $\bm{Q}^{s}$ of the current SSCA module:
\begin{align}
    \bm{Q}^{s} \coloneqq \bm{F}^{s-1},
\end{align}
where $s = 1, 2, \dots, S$ and $S$ is the number of cascaded SSCA modules. The $\bm{Q}^{0}$ is set by Kaiming initialization \cite{he2015delving}.

The extracted label features $\bm{F}^{s}$ in each SSCA module are sent to the independent classifiers, and the predicted logits are averaged as the final prediction. In order to be in line with the second observation in Sec. \ref{sec:two_observ} and hypothesis made in Eq. \ref{eq:optim_theta}, \ie, the norms of classifiers are equal, we fix the individual classifier weight $\| w_j \| = 1$  by L2 normalization for each attribute. We also fix the label feature $\| \bm{F}_{i} \|$ by L2 normalization and re-scale it to $\gamma = 30$ by default. The normalization of classifier weights and label features makes the model focus on the learning of angles, which is the decisive factor in predictions, as pointed out in Eq. \ref{eq:angle}.

Following the common practice, we adopt the binary cross-entropy loss as the classification loss:
\begin{align}
    L_{cls} & = \frac{1}{N} \sum_{i=1}^{N} \sum_{j=1}^{M} (y_{i,j} \log(p_{i,j})  + (1 - y_{i,j})\log(1 - p_{i,j}))
\end{align}
where $p_{i,j} = \sigma( \gamma \cdot \cos{\theta})$ and $\cos{\theta} = \frac{w^{T}_{j} \bm{F_{i, j}}}{\|w_{j}\|  \cdot \|\bm{F_{i, j}}\|}$.
The total loss is formulated as $L = \sum^{S}_{s=1} L^{s}_{cls}$, where $S$ is the number of the SSCA modules.


\section{Experiments} \label{sec:experiments}


In this section, we conduct experiments on eight datasets of three tasks, \ie, generic multi-label classification, pedestrian attribute recognition, and continual multi-label learning, to validate the effectiveness of our method. 

\subsection{Generic Multi-Label Classification}
As the most common multi-label related task, the generic multi-label classification aims to assign multiple categories to natural images, as shown in Fig. \ref{fig:task_demo}(a). Experiments are conducted on MS-COCO \cite{lin2014microsoft}, PASCAL-VOC \cite{everingham2010pascal}, and NUS-WIDE \cite{chua2009nus} datasets to evaluate our method. We first introduce the dataset details and evaluation metrics. Then, performance comparisons with state-of-the-arts are given to show the superiority of our method. Finally, we conduct ablation studies on MS-COCO datasets to verify the effects of the SSCA module in our DLFL framework.

\subsubsection{Datasets}

\textbf{MS-COCO} \cite{lin2014microsoft} dataset (version published in 2014)  is the most prevalent benchmark for evaluating multi-label image classification algorithms. Following the standard settings in previous works \cite{chen2021P_GCN, ridnik2021asymmetric, zhao2021transformer}, we adopt 82,081 images as the training set and 40,504 images as the test set. Natural scene images in the training and test sets are labeled with 80 categories, such as person, car, broccoli, and toothbrush. Due to the large size and rich annotated labels, MS-COCO is one of the most challenging datasets in the generic multi-label classification task.

\textbf{PASCAL-VOC} \cite{everingham2010pascal} dataset proposed in the PASCAL Visual Object Classes Challenge 2007 is another widely used benchmark for generic multi-label classification. It comprises 5011 training and 4952 test images, labeled with 20 categories. 

\textbf{NUS-WIDE} \cite{chua2009nus} dataset is collected from the Flickr website. Following the original settings and previous work \cite{chen2021P_GCN}, we adopt the protocol with 161,789 training images and 107,859 test images. Each image is annotated with 81 labels.

\subsubsection{Evaluation Metrics}

Following the previous work settings \cite{chen2021P_GCN, ridnik2021asymmetric, zhao2021transformer, zhang2013review}, three types of metrics are adopted to evaluate the model performance. The first type of metrics is the average precision (AP) of each class and the mean average precision (mAP) over all classes, which are computed as follows \cite{song2018deep, chen2021P_GCN}:
\begin{align}
    AP(y_{j}) & = \frac{1}{N_{j}} \sum_{n=1}^{\hat{N}_{j}} Prec_{y_{j}}(n) \times (Rec_{y_{j}}(n) - Rec_{y_{j}}(n-1)) \nonumber \\ 
    mAP & = \frac{1}{M} \sum_{m=1}^{M} AP(y_{j}) \nonumber
\end{align}
where $N_{j}$ is the number of the images relevant to the $j$-th label, $\hat{N}_{j}$ is the total number of retrieved images for the $j$-th label, $n$ is the rank in the list of retrieved images, and $M$ is the number of labels. Precision and recall at the rank $n$ is denoted as $Prec_{y_{j}}(n)$ and $Rec_{y_{j}}(n)$. The second type of metrics is the macro-average metric, \ie, per-class precision (CP), per-class recall (CR), and per-class F1 (CF1), which compute the performance independently for each class and then take the average. The third type of metric is the micro-average metric, \ie, overall precision (OP), overall recall (OR), and overall F1 (OF1), which compute the performance for all labels in a holistic view without differentiating categories. The macro-average and micro-average metrics are given as follows:
\begin{align}
    & CP = \frac{1}{M} \sum_{j=1}^{M} \frac{TP_{j}}{TP_{j} + FP_{j}}, &
    & OP = \frac{\sum_{m=1}^{M} TP_{j}}{\sum_{j=1}^{M} (TP_{j} + FP_{j})},  \nonumber\\
    & CR = \frac{1}{M} \sum_{j=1}^{M} \frac{TP_{j}}{TP_{j} + FN_{j}},  &
    & OR = \frac{\sum_{m=1}^{M} TP_{j}}{\sum_{j=1}^{M} (TP_{j} + FN_{j})}, \nonumber\\
    & CF1 = \frac{2 * CP * CR}{CP + CR}, &
    & OF1 = \frac{2 * OP * OR}{OP + OR} \nonumber
\end{align}
where $TP_{j}$, $FP_{j}$, and $FN_{j}$ are the number of true positive, false positive, and false negative samples of label $j$.
For performance on MS-COCO \cite{lin2014microsoft} and NUS-WIDE \cite{chua2009nus} datasets, all three types of metrics are adopted to evaluate the effect of methods. For performance on PASCAL-VOC \cite{everingham2010pascal} datasets, only AP and mAP are adopted.

\begin{table*}[t]
\centering{
\setlength\tabcolsep{4pt}
\caption{\textbf{Performance comparisons with state-of-the-art methods on the MS-COCO dataset}. We report the  performance of our approach on two resolution settings (448, 448), (576, 576) and two backbone networks, ResNet-101 \cite{he2016deep} and Swin Transformer \cite{liu2021swin}. $R_{train}$ and $R_{test}$ denote the resolution used in the training and test stage. The \textbf{first} highest scores are represented by bold font.}
\label{table: COCO benchmark}
\renewcommand{\arraystretch}{1.2}
\begin{tabular}{c|c|c|c|cccccc|cccccc}
\toprule
\multirow{2}{*}{Methods} & \multirow{2}{*}{Backbone} & \multirow{2}{*}{$(R_{train}, R_{test})$} & \multirow{2}{*}{mAP} & \multicolumn{6}{c|}{All}  & \multicolumn{6}{c}{Top 3} \\ \cline{5-16} 
&    &  &  & CP   & CR   & CF1  & OP   & OR   & OF1  & CP   & CR   & CF1  & OP   & OR   & OF1  \\ \hline \hline
CNN-RNN \cite{wang2016cnn} & VGG + LSTM &  --  & 61.2 & -- & -- & -- & -- & -- & -- & 66.0 & 55.6 & 60.4 & 69.2 & 66.4 & 67.8 \\
RNN-Attention \cite{wang2017multi}  & VGG + LSTM &  -- & -- & -- & -- & -- & -- & -- & --  & 78.0 & 57.7 & 66.3 & 83.8 & 62.3 & 71.4\\
Order-Free RNN \cite{chen2018order} & Resnet152 + LSTM &  -- & -- & -- & -- & -- & -- & -- & --   & 71.6 & 54.8 & 62.1 & 74.2 & 62.2 & 67.7 \\
SRN \cite{zhu2017learning}  & ResNet-101 & (224, 224)  & 77.1 & 81.6 & 65.4 & 71.2 & 82.7 & 69.9 & 75.8 & 85.2 & 58.8 & 67.4 & 87.4 & 62.5 & 72.9 \\
PLA  \cite{yazici2020orderless} & ResNet-101 + LSTM & (288, 288) & --& 80.4 & 68.9 & 74.2 & 81.5 & 73.3 & 77.2 & -- & -- & -- & -- & -- & --  \\

ML-GCN \cite{chen2019multi} & ResNet-101 & (448, 448) & 83.0 & 85.1 & 72.0 & 78.0 & 85.8 & 75.4 & 80.3 & 89.2 & 64.1 & 74.6 & 90.5 & 66.5 & 76.7 \\
KSSNet \cite{wang2020multi}  & ResNet-101 & (448, 448) & 83.7 & 84.6 & 73.2 & 77.2 & 87.8 & 76.2 & 81.5 & -- & -- & -- & -- & -- & -- \\
P-GCN \cite{chen2021learning} & ResNet-101 & (448, 448) & 83.2 & 84.9 & 72.7 & 78.3 & 85.0 & 76.4 & 80.5 & 89.2 & 64.3 & 74.8 & 90.0 & 66.8 & 76.7\\
CSRA \cite{zhu2021residual} & ResNet-101 & (448, 448) & 83.5 & 84.1 & 72.5 & 77.9 & 85.6 & 75.7 & 80.3 & 88.5 & 64.2 & 74.4 & 90.4 & 66.4 & 76.5 \\
TDRG \cite{zhao2021transformer} & ResNet-101 & (448, 448) & 84.6  & 86.0 & 73.1 & 79.0 & 86.6 & 76.4 & 81.2 & 89.9 & 64.4 & 75.0 & 91.2 & 67.0 & 77.2 \\
ADD-GCN \cite{ye2020attention} & ResNet-101 & (448, 448) & 84.2 & 85.1 & 73.6 & 79.0 & 86.2 & 76.6 & 81.1 & 89.2 & 65.2 & 75.3 & 90.9 & 67.1 & 77.2 \\ 
MCAR \cite{gao2021MCAR} & ResNet-101 & (448, 448) & 83.8 & 85.0 & 72.1 & 78.0 & 88.0 & 73.9 & 80.3 & 88.1 & 65.5 & 75.1 & 91.0 & 66.3 & 76.7\\ 
SST \cite{chen2022sst} &  ResNet-101 & (448, 448) & 84.2 & 86.1 & 72.1 & 78.5 & 87.2 & 75.4 & 80.8 & 89.8 & 64.1 & 74.8 & 91.5 & 66.4 & 76.9 \\ \hline \hline
baseline & ResNet-101 & (448, 448) & 81.4 & 83.7 & 70.5 & 76.5 & 85.8 & 74.1 & 79.5 &  86.4 & 62.9 & 72.8 & 90.5 & 65.6 & 76.0\\
baseline & ResNet-101 & (448, 640) & 82.0 & 90.6 & 60.4 & 72.5 & 93.9 & 64.9 & 76.8 & 91.8 & 56.6 & 70.0 & 95.1 & 60.5 & 74.0 \\
baseline & Swin-B & (448, 448) & 85.8 & 87.0 & 74.7 & 80.4 & 88.1 & 77.6 & 82.5 & 90.4 & 66.0 & 76.3 & 92.3 & 68.1 & 78.4 \\
baseline & Swin-B & (448, 640) & 85.9 & \bf{94.4} & 63.4 & 75.9 & \bf{95.1} & 67.2 & 78.7 & \bf{95.5} & 59.5 & 73.3 & \bf{96.2} & 62.8 & 75.9 \\
Ours & ResNet-101 & (448, 448) & 84.4 & 85.5 & 73.9 & 79.2 & 86.4 & 76.9 & 81.3 & 89.2 & 65.1 & 75.3 & 91.0 & 67.2 & 77.3\\
Ours & ResNet-101 & (448, 640) & 85.9 & 87.1 & 75.1 & 80.7 & 88.2 & 77.8 & 82.7 & 90.3 & 66.2 & 76.4 & 92.1 & 67.9 & 78.2  \\
Ours & Swin-B & (448, 448) & 88.3 & 88.3 & 78.0 & 82.8 & 89.1 & 80.0 & 84.3 & 91.8 & 68.2 & 78.3 & 93.3 & 69.5 & 79.7 \\
Ours & Swin-B & (448, 640) & 89.4 & 90.3 & 78.0 & 83.7 & 91.0 & 79.8 & 85.0 & 93.1 & 68.5 & 78.9 & 94.4 & 69.5 & 80.1 \\
\hline \hline
SSGRL \cite{chen2019learning} &  ResNet-101 & (576,576) & 83.8 & 89.9 & 68.5 & 76.8 & 91.3 & 70.8 & 79.7 & 91.9 & 62.5 & 72.7 & 93.8 & 64.1 & 76.2 \\
C-Tran \cite{lanchantin2021CTran} & ResNet-101 & (576,576) & 85.1 & 86.3 & 74.3 & 79.9 & 87.7 & 76.5 & 81.7 & 90.1 & 65.7 & 76.0 & 92.1 & \bf{71.4} & 77.6 \\
TDRG \cite{zhao2021transformer} & ResNet-101 & (576,576) & 86.0 & 87.0 & 74.7 & 80.4 & 87.5 & 77.9 & 82.4 & 90.7 & 65.6 & 76.2 & 91.9 & 68.0 & 78.1 \\ 
MCAR \cite{gao2021MCAR} & ResNet-101 & (576,576) & 84.5 & 84.3 & 73.9 & 78.7 & 86.9 & 76.1 & 81.1 & 87.8 & 65.9 & 75.3 & 90.4 & 67.1 & 77.0 \\
\hline
Ours & ResNet-101 & (576,576) & 86.0 & 86.8 & 75.3 & 80.6 & 87.5 & 78.1 & 82.6 & 90.6 & 66.2 & 76.5 & 92.1 & 68.0 & 78.2 \\
Ours & ResNet-101 & (576,640) & 86.3 & 86.9 & 75.8 & 81.0 & 87.7 & 78.7 & 83.0 & 90.5 & 66.3 & 76.5 & 92.2 & 68.3 & 78.5 \\
Ours & Swin-B & (576,576) & 89.2 & 88.8 & 79.2 & 83.8  & 89.6 & 81.0 & 85.1 & 92.3 & 69.1 & 79.0 & 93.6 & 70.0 & 80.1 \\  
Ours & Swin-B & (576,640) & \bf{89.6} & 88.9 & \bf{79.9} & \bf{84.2} & 89.6 & \bf{81.8} & \bf{85.5} & 92.4 & \bf{69.2} & \bf{79.2} & 93.8 & 70.3 & \bf{80.4} \\
\bottomrule
\end{tabular}
}
\end{table*}

\subsubsection{Experimental settings}
To make a fair comparison, we adopt ResNet-101 \cite{he2016deep} and Swin Transfomer \cite{liu2021swin} pre-trained on the ImageNet \cite{deng2009imagenet} as the backbone network respectively. Following the common setting, training images of generic multi-label classification datasets are resized to (448, 448) and (576, 576) as the inputs, respectively. Multi-scale cropping and horizontal flipping are used as the data augmentations. We adopt the SGD optimizer with 1e-4 weight decay to train our model. The initial learning rate of the backbone network and classification head is set to 5e-4 and 5e-3 separately. We train the DLFL framework 20 epochs with the batch size 32.

\subsubsection{Comparison with SOTA Methods}

In this section, we compare our proposed method with previous methods in terms of performance and complexity.  For performance comparison, we conduct experiments on MS-COCO \cite{lin2014microsoft}, PASCAL-VOC \cite{everingham2010pascal}, and NUS-WIDE \cite{chua2009nus} datasets following the common settings to make a fair comparison. For complexity comparison, we focus on the methods based on the OFOL mechanism and report the parameters and computations to measure the model complexity.

\textbf{Performance on MS-COCO.} Comparison on MS-COCO \cite{lin2014microsoft} with state-of-the-art (SOTA) methods are given in Tab. \ref{table: COCO benchmark}. We compare our approach with as many SOTA methods as possible. First, compared to the baseline methods with the ResNet-101 \cite{he2016deep} and Swin-B \cite{liu2021swin} backbones, our approach achieves a performance improvement of 3.0 on ResNet-101 and 2.5 on Swin-B. The consistent performance improvement in different backbones demonstrates the effectiveness of our SSCA module as a feature disentangle module independent from the image feature extraction (backbone) network. Second, compared to SOTA method TDRG, our method with ResNet-101 achieves comparable performance on the same experimental settings. Combined with the Swin-B backbone network, our method achieves state-of-the-art performance 88.3 and 89.2 mAP in (448, 448) and (576, 576) input sizes, respectively. We also find that increasing the input size of the inference stage can further improve the performance of our method. Increasing the inference image size of the baseline method with ResNet-101 and Swin-b only brings a performance improvement of 0.6 and 0.4 in mAP. Compared with the baseline method, our method improves the mAP performance from 84.4 and 88.3 to 85.9 (+1.5) and 89.4 (+1.1) on ResNet-101 and Swin-b backbones. This phenomenon is attributed to the design of our SSCA module, which retrieves and aggregates label-related features from the image feature map by semantic queries. Therefore, increasing the input image size enables the semantic queries to retrieve more accurate and fine-grained label-related features. Finally, our method can achieve 89.6, 84.2, and 85.5 in mAP, CF1, and OF1 metrics with the Swin-B backbone network, training image size 576, and test image size 640.

\begin{table*}[t]
\centering{
\setlength\tabcolsep{3pt}
\caption{\textbf{Performance comparisons with state-of-the-art methods on the PASCAL-VOC dataset}. The performance of our approach based on ResNet-101 and resolution $448\times 448$ is reported. We update the SOTA performance in 15 of the 20 classes. }
\label{table: VOC2007 benchmark}
\resizebox{\textwidth}{!}{
\begin{tabular}{c|c|c|c|c|c|c|c|c|c|c|c|c|c|c|c|c|c|c|c|c||c}
\toprule
Methods & aero & bike & bird & boat & bottle & bus & car & cat & chair & cow & table & dog & horse & motor & person & plant & sheep & sofa & train & tv & mAP \\ \hline \hline
CNN-RNN \cite{wang2016cnn} & 96.7 & 83.1 & 94.2 & 92.8 & 61.2 & 82.1 & 89.1 & 94.2 & 64.2 & 83.6 & 70.0 & 92.4 & 91.7 & 84.2 & 93.7 & 59.8 & 93.2 & 75.3 & 99.7 & 78.6 & 84.0 \\
RNN-Attention \cite{wang2017multi} & 98.6 & 97.4 & 96.3 & 96.2 & 75.2 & 92.4 & 96.5 & 97.1 & 76.5 & 92.0 & 87.7 & 96.8 & 97.5 & 93.8 & 98.5 & 81.6 & 93.7 & 82.8 & 98.6 & 89.3 & 91.9 \\
Fev+Lv \cite{yang2016exploit} & 97.9 & 97.0 & 96.6 & 94.6 & 73.6 & 93.9 & 96.5 & 95.5 & 73.7 & 90.3 & 82.8 & 95.4 & 97.7 & 95.9 & 98.6 & 77.6 & 88.7 & 78.0 & 98.3 & 89.0 & 90.6 \\
Atten-Reinforce \cite{chen2018recurrent} & 98.6 & 97.1 & 97.1 & 95.5 & 75.6 & 92.8 & 96.8 & 97.3 & 78.3 & 92.2 & 87.6 & 96.9 & 96.5 & 93.6 & 98.5 & 81.6 & 93.1 & 83.2 & 98.5 & 89.3 & 92.0 \\
SSGRL \cite{chen2019learning}  & 99.5 & 97.1 & 97.6 & 97.8 & 82.6 & 94.8 & 96.7 & 98.1 & 78.0 & 97.0 & 85.6 & 97.8 & 98.3 & 96.4 & 98.1 & 84.9 & 96.5 & 79.8 & 98.4 & 92.8 & 93.4 \\
ML-GCN \cite{chen2019multi} & 99.5 & 98.5 & 98.6 & 98.1 & 80.8 & 94.6 & 97.2 & 98.2 & 82.3 & 95.7 & 86.4 & 98.2 & 98.4 & 96.7 & 99.0 & 84.7 & 96.7 & 84.3 & 98.9 & 93.7 & 94.0 \\
ADD-GCN \cite{ye2020attention} & 99.7 & 98.5 & 97.6 & 98.4 & 80.6 & 94.1 & 96.6 & 98.1 & 80.4 & 94.9 & 85.7 & 97.9 & 97.9 & 96.4 & 99.0 & 80.2 & 97.3 & 85.3 & 98.9 & 94.1 & 93.6 \\ 
TDRG \cite{zhao2021transformer} & 99.9 & 98.9 & 98.4 & 98.7 & 81.9 & 95.8 & 97.8 & 98.0 & \textbf{85.2} & 95.6 & \textbf{89.5} & 98.8 & 98.6 & 97.1 & 99.1 & 86.2 & 97.7 & \textbf{87.2} & 99.1 & \textbf{95.3} & 95.0 \\ 
P-GCN \cite{chen2021learning} & 99.6 & 98.6 & 98.4 & \textbf{98.7} & 81.5 & 94.8 & 97.6 & 98.2 & 83.1 & 96.0 & 87.1 & 98.3 & 98.5 & 96.3 & 99.1 & 87.3 & 95.5 & 85.4 & 98.9 & 93.6 & 94.3 \\ \hline \hline
Baseline & 99.8 & 97.7 & 97.8 & 97.9 & 80.9 & 93.4 & 96.6 & 97.6 & 77.6 & 93.7 & 86.4 & 97.7 & 97.8 & 96.0 & 98.9 & 81.1 & 95.8 & 80.1 & 99.1 & 90.4 & 92.8\\
Ours & \textbf{100.0} & \textbf{98.9} & \textbf{98.8} & 98.2 & \textbf{87.9} & \textbf{98.3} & \textbf{98.1} & \textbf{98.9} & 82.1 & \textbf{98.5} & 87.4 & \textbf{99.2} & \textbf{99.4} & \textbf{97.9} & \textbf{99.2} & \textbf{88.9} & \textbf{99.1} & 85.3 & \textbf{99.7} & 93.4 & \textbf{95.4} \\  \bottomrule
\end{tabular}
}
}
\end{table*}

\begin{table*}[t]
\centering{
\setlength\tabcolsep{4pt}
\caption{\textbf{Performance comparisons with state-of-the-art methods on the NUS-WIDE dataset}. We conduct experiments based on ResNet-101, and input image size is resized to $448\times 448$.}
\label{table: NUS benchmark}
\renewcommand{\arraystretch}{1.2}
\begin{tabular}{c|c|cccccc|cccccc}
\toprule
\multirow{2}{*}{Methods} & \multirow{2}{*}{mAP} & \multicolumn{6}{c|}{All}  & \multicolumn{6}{c}{Top 3} \\ \cline{3-14} 
&  & CP   & CR   & CF1  & OP   & OR   & OF1  & CP   & CR   & CF1  & OP   & OR   & OF1  \\ \hline \hline
kNN \cite{chua2009nus}  & -- & -- & -- & -- & -- & -- & -- & 32.6 & 19.3 & 24.3 & 42.9 & 53.4 & 47.6 \\
WARP \cite{gong2013deep}  & -- & -- & -- & -- & -- & -- & --  & 31.7 & 35.6 & 33.5 & 48.6 & 60.5 & 53.9\\
CNN-RNN \cite{wang2016cnn} & -- & -- & -- & -- & -- & -- & --   & 40.5 & 30.4 & 34.7 & 49.9 & 61.7 & 55.2 \\
ML-ZSL \cite{lee2018multi} & -- & -- & -- & -- & -- & -- & -- & 43.4 & 48.2 & 45.7 & -- & -- & -- \\
SRN  \cite{zhu2017learning} & 62.0 & 65.2 & 55.8 & 58.5 & 75.5 & 71.5 & 73.4 & 48.2 & \textbf{58.8} & 48.9 & 56.2 & \textbf{69.6} & 62.2  \\
ResNet-101 \cite{chen2021learning} & 59.7 & 62.6 & 53.3 & 57.5 & 75.4 & 70.2 & 72.7 & 63.8 & 46.6 & 53.9 & 77.9 & 60.9 & 68.4 \\
C-GCN \cite{chen2019multi}  & 62.5 & 66.9 & 54.5 & 60.1 & \textbf{76.4} & 70.2 & 73.1 & 67.8 & 49.2 & 57.0 & 78.1 & 61.7 & 68.9 \\
P-GCN \cite{chen2021learning}  & 62.8 & 64.4 & 56.8 & 60.4 & 75.7 & 71.2 & 73.4 & 66.4 & 49.9 & 57.0 & 78.4 & 61.8 & 69.1\\\hline \hline
Baseline & 61.2 & 65.7 & 52.9 & 58.6 & 76.7 & 67.8 & 72.0 & 67.3 & 47.0 & 55.4 & 78.8 & 59.7 & 68.0 \\
Ours & \textbf{63.7} & \textbf{67.0} & \textbf{56.8} & \textbf{61.5} & 76.0 & \textbf{73.2} & \textbf{74.5} & \textbf{68.7} & 48.7 & \textbf{57.0} & \textbf{78.9} & 63.1 & \textbf{70.1}\\
\bottomrule
\end{tabular}
}
\end{table*}

\textbf{Performance on PASCAL-VOC.} Quantitative results on PASCAL-VOC are shown in Tab. \ref{table: VOC2007 benchmark}. We update the best performance of 15 out of 20 classes on the PASCAL-VOC \cite{everingham2010pascal} dataset and achieve 95.4 in mAP. Compare to SOTA method TDRG \cite{zhao2021transformer}, we achieve 0.4 performance improvement in mAP. Since the performance of SOTA method on PASCAL-VOC dataset is close to 95, the performance improvement of 0.4 is nontrivial.

\textbf{Performance on NUS-WIDE.} Since the noisy labels and low-resolution images, NUS-WIDE dataset is the one of the most challenging dataset in generic multi-label classification. As shown in Tab. \ref{table: NUS benchmark}, our method achieve 63.7 performance in mAP, which is 0.9 higher than the P-GCN \cite{chen2021P_GCN}.

\textbf{Complexity Analysis.} We evaluate the model complexity from the perspectives of parameters and computations. As presented in Tab. \ref{tab:complexity}, we compare our method with TDRG \cite{zhao2021transformer}, ADD-GCN \cite{ye2020attention}, SSGRL \cite{chen2019learning}, and C-TRANS \cite{lanchantin2021CTran}, which follow the OFOL mechanism and try to extract a specific feature for each labels. It is obvious that our method achieves comparable or even better performance with less complexity. Compared with TDRG \cite{zhao2021transformer}, our approach achieves the same performance in mAP and better performance in CF1 and OF1 with 83.9\% of the computational cost and 87.2\% of the parameters.

\begin{table}[h]
\centering
\caption{\textbf{Complexity comparison with state-of-the-art methods following the OFOL mechanism}. The \textbf{first} and \underline{second} highest scores are represented by bold font and underline respectively.} 
\resizebox{\linewidth}{!}{
\begin{tabular}{c|c|c|c|c|c|c}
\toprule
	Methods & Resolution & mAP & CF1 & OF1 & Params(M) & Flops(G) \\\midrule \midrule
	ADD-GCN \cite{ye2020attention} & 448 & 84.2 & 79.0 & 81.1 & \bf{48.2} & \bf{32.0} \\
	SSGRL \cite{chen2019learning} & 576 & 83.8 & 76.8 & 79.9 & 92.3 & 89.5 \\
	TDRG \cite{zhao2021transformer} & 576 & \bf{86.0} & \underline{80.4} & \underline{82.4} & 68.3 & 64.7 \\
	C-TRANS \cite{lanchantin2021CTran} & 576 & 85.1 & 79.9 & 81.7 & 118.4 & 62.0 \\ \hline \hline
	Ours & 576 & \bf{86.0} & \bf{80.6} & \bf{82.6} & \underline{59.6} & \underline{54.3} \\
\bottomrule
\end{tabular}}
\label{tab:complexity}
\end{table}

\subsubsection{Ablation Study}

In this section, we conduct experiments to investigate the effect of the number of the SSCA modules. 
As shown in Tab. \ref{tab:ssca_number}, as the number of SSCA module $S$ increases, the performance rises slightly from 84.44 to 84.54. We argue that the cascaded SSCA modules are beneficial for discriminative semantic queries and accurate query attention map when $S$ increases. 
However, the increased number of SSCA modules brings more computation costs and complexity to the model. Our work aims to provide a straightforward implementation of the feature disentangling module in Fig.\ref{fig:OFML_OFOL} to validate the superiority of the OFML mechanism. Thus, we choose the $S = 1$ as our method by default for generic multi-label classification.

\begin{table}[t]
\centering
\caption{{\bf Experiments on the number of cascaded SSCA modules}. $S = 0$ indicates the baseline model without the SSCA module.} 
\begin{tabular}{c|c|c|c|c|c}
\toprule
	Number & mAP & CF1 & OF1 & Params(M) & Flops(G)\\\midrule \midrule
	$S = 0$ & 81.39 & 76.46 & 79.54 & 42.7 & 31.4 \\
	$S = 1$ & 84.44 & 79.23 & 81.34 & 59.6 & 33.2 \\
	$S = 2$ & 84.38 & 79.10 & 81.37 & 72.4 & 34.7\\
	$S = 3$ & \bf{84.54} & \bf{79.32} & \bf{81.40} & 85.1 & 36.2\\
	$S = 4$ & 84.42 & 78.94 & 81.26 & 97.9 & 37.7\\
	$S = 5$ & 84.52 & 79.05 & 81.28 & 110.7 & 39.2\\
\bottomrule
\end{tabular}
\label{tab:ssca_number}
\end{table}

In addition, our method performs best for pedestrian attribute recognition when the number of SSCA modules is set to $S = 3$, verified in our conference paper \cite{jia2022learning}. We attribute the difference in the optimal number of SSCA modules on the two tasks to the different image resolutions. For generic multi-label classification, natural scene images of datasets have a large size and are resized to (448, 448) as inputs. On the contrary, for pedestrian attribute recognition, cropped pedestrian images from surveillance video frames have small resolutions and are resized to (256, 192) as inputs. One SSCA module can extract representative label features from fine-grained image feature maps when input images are large. Instead, since the images of pedestrian attribute datasets are blurry, more SSCA modules are required to refine the label features iteratively.

\subsection{Pedestrian attribute recognition}

\begin{figure}[t]
	\centering
	\subfloat[Samples of pedestrian attribute datasets.]{
		\includegraphics[width=1\linewidth ]{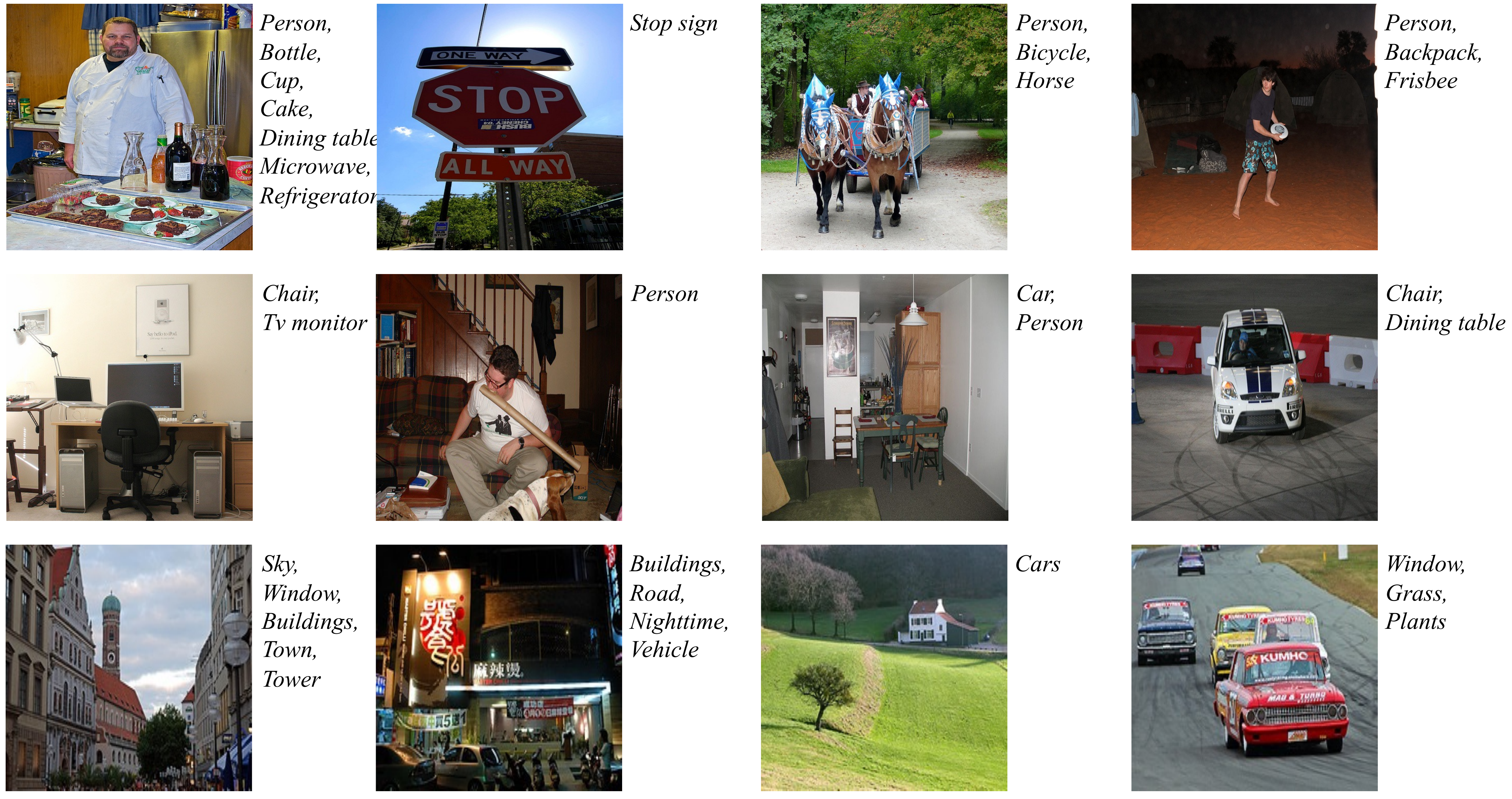}}  \\
	\subfloat[Samples of generic multi-label datasets and continual multi-label learning datasets.]{
		\includegraphics[width=1\linewidth ]{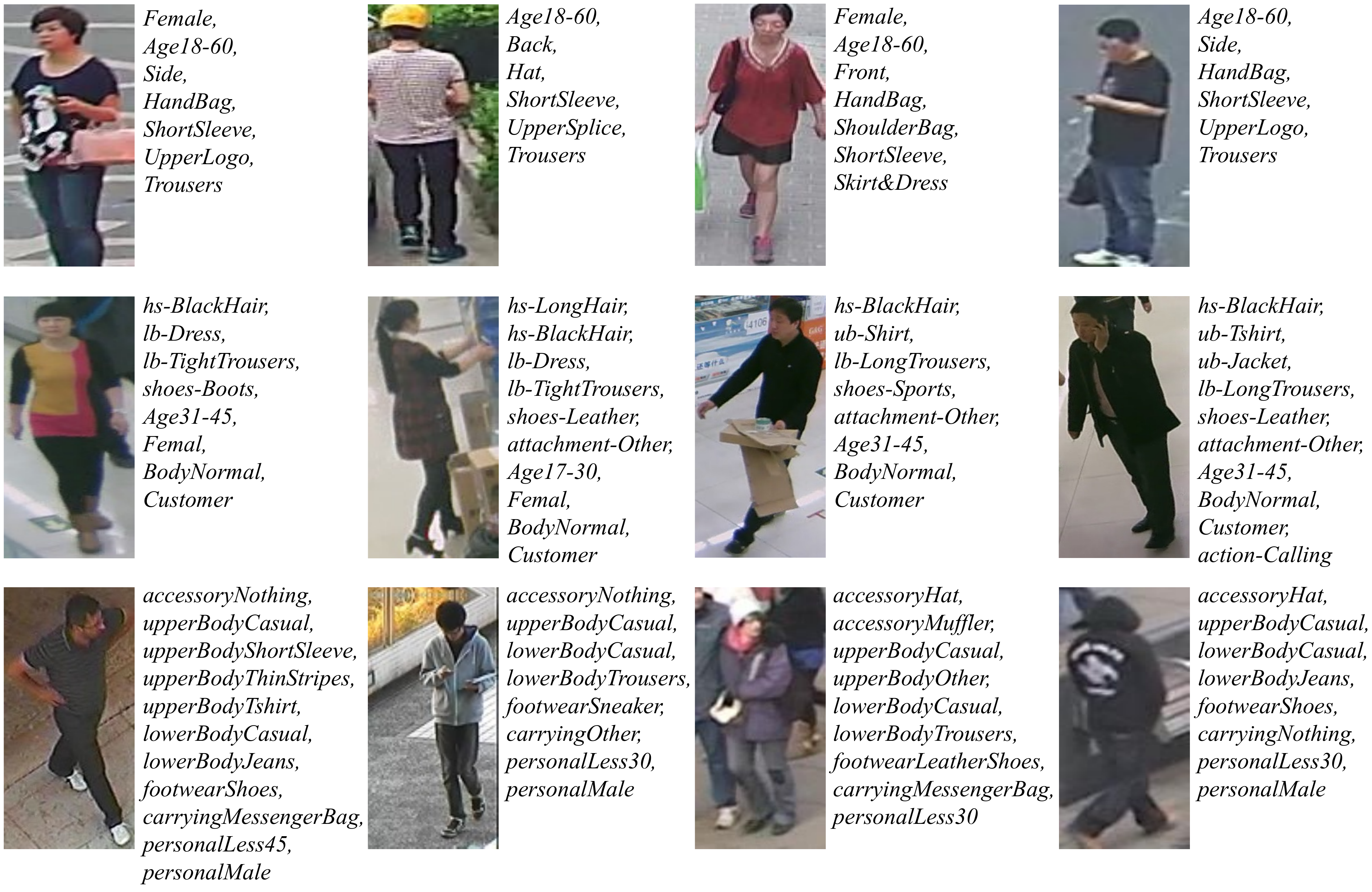}}
	\caption{\textbf{Illustration of images and labels of three tasks}. (a) Samples of PA-100K, $\text{RAP}_{\text{ZS}}$, and $\text{PETA}_{\text{ZS}}$ are listed in the first, second, and third rows. (b) Samples of MS-COCO, PASCAL-VOC, and NUS-WIDE are listed in the first, second, and third rows. }
	\label{fig:task_demo}
\end{figure}

As the application of multi-label classification in surveillance scenarios, pedestrian attribute recognition aims to predict multiple human attributes for a pedestrian image captured by surveillance cameras. These human attributes include gender, age, clothes, and accessories. Considering the essential requirement of practical industry application, we evaluate our method on PA-100K, $\text{RAP}_{\text{ZS}}$, and $\text{PETA}_{\text{ZS}}$ following the zero-shot pedestrian identity settings \cite{jia2021rethinking}. Pedestrian images are listed in Fig. \ref{fig:task_demo} (b). Compared to images of generic multi-label datasets, images in pedestrian attribute datasets are smaller and blurry with strongly changing illumination. In addition, unlike the significant semantic differences between object categories in the generic multi-label datasets (``person" \vs ``cat"), the semantic differences between pedestrian attributes are much more minor ("UpperStride" \vs ``UpperLogo"). 

\subsubsection{Datasets}

\textbf{PA-100K} dataset \cite{liu2017hydraplus} is the largest pedestrian attribute dataset collected from outdoor surveillance cameras. It consists of 100,000 pedestrian images and is divided into training, validation, and test sets with a ratio of 8:1:1. Each image is described with 26 commonly used human attributes.

\textbf{$\text{RAP}_{\text{ZS}}$} \cite{jia2021rethinking} dataset is collected from a realistic surveillance network at an indoor shopping mall. According to the zero-shot identity setting, 26,632 images of 2,590 pedestrian identities are divided into 17,062 images of 1,566 identities in the training set, 4,648 images of 505 identities in the validation set, and 4,928 images of 518 identities in the test set. Each image is labeled with 69 binary attributes and 3 multi-class attributes. Due to the small sample size of attributes, we follow the protocol \cite{li2018richly, jia2021rethinking} and adopt 53 binary attributes to evaluate the recognition performance.

\textbf{$\text{PETA}_{\text{ZS}}$} \cite{jia2021rethinking} dataset is reconstructed from the PETA \cite{deng2014pedestrian}, which is collected from 10 small-scale person datasets and contains 19,000 images of 8,699 pedestrian identities. Each image is labeled with 61 binary attributes and 4 multi-class attributes. We follow the common experimental protocol \cite{deng2014pedestrian, tang2019Improving, jia2021spatial, jia2022learning}, and only 35 attributes whose positive ratios are higher than 5\% are used for evaluation. According to the zero-shot identity setting, all images are divided into 11,241 training images of 5,233 identities, 3,826 validation images of 1,760 identities, and 3,933 test images of 1,706 identities.

\subsubsection{Evaluation Metrics}

\begin{table*}[t]
\centering{
\setlength\tabcolsep{4pt}
\caption{\textbf{Comparison with state-of-the-art methods on the $\text{PETA}_{\text{ZS}}$, $\text{RAP}_{\text{ZS}}$, and PA-100K datasets}. Performance in five metrics, including mean accuracy (mA), accuracy (Accu), precision (Prec), recall, and F1, is evaluated. The first and \underline{second} highest scores are represented by bold font and underline, respectively.}
\label{table: pedes benchmark}
\renewcommand{\arraystretch}{1.2}
\begin{tabular}{c|c|ccccc|ccccc|ccccc}
\toprule
\multirow{2}{*}{Method} & \multirow{2}{*}{Backbone} & \multicolumn{5}{c|}{$\text{PETA}_\text{ZS}$} & \multicolumn{5}{c|}{$\text{RAP}_\text{ZS}$} &\multicolumn{5}{c}{PA-100K} \\ \cline{3-17} 
	&	& mA & Accu & Prec & Recall & F1 & mA & Accu & Prec & Recall & F1 & mA & Accu & Prec & Recall & F1\\ \hline \hline
MsVAA \cite{sarafianos2018deep} & ResNet-101 & 71.53 & 58.67 & 74.65 & 69.42 & 71.94 & 72.04 & 62.13 & 75.67 & 75.81 & 75.74 & -- & -- & -- & -- & -- \\
VRKD \cite{li2019visual} & ResNet-50 & -- & -- & -- & -- & -- & -- & -- & -- & -- & -- & 77.87 & 78.49 & 88.42 & 86.08 & 87.24 \\
VAC  \cite{guo2019visual} & ResNet-50 & 71.91 & 57.72 & 72.05 & 70.64 & 70.90 & 73.70 & 63.25 & 76.23 & 76.97 & 76.12 & 79.16 & 79.44 & \bf{88.97} & 86.26 & 87.59 \\
ALM \cite{tang2019Improving} & BN-Inception & 73.01 & 57.78 & 69.50 & \bf{73.69} & 71.53 & 74.28 & 63.22 & 72.96 & 80.73 & 76.65 & 80.68 & 77.08 & 84.21 & 88.84 & 86.46 \\ 
JLAC \cite{tan2020relation} & ResNet-50 & 73.60 & 58.66 & 71.70 & 72.41 & 72.05 & 76.38 & 62.58 & 73.14 & 79.20 & 76.05 & 82.31 & 79.47 & 87.45 & 87.77 & 87.61 \\ 
SSC \cite{jia2021spatial} & ResNet-50 & 70.71 & 58.42 & \bf{74.96} & 69.00 & 71.31 & 72.20 & 65.25 & \bf{78.78} & 77.17 & 77.56 & 81.70 & 78.85 & 85.80 & 88.92 & 86.89 \\ \hline \hline
Baseline & ResNet-50 & 71.43 & 58.69 & 74.41 & 69.82 & 72.04 & 71.76 & 64.83 & 78.75 & 76.60 & 77.66 &  80.71 & 78.61 & 86.50 & 87.59 & 87.04 \\
Ours & ResNet-50 & \bf{75.52} & \bf{58.83} & 70.72 & 73.65 & \bf{72.16} & \bf{77.83} & \bf{64.86} & 74.46 & \bf{81.55} & \bf{77.85} & \bf{84.68} & \bf{80.23} & 86.72 & \bf{89.44} & \bf{88.16}\\ \bottomrule
\end{tabular}
}
\end{table*}

Two types of metrics, \ie, a label-based metric and four instance-based metrics, are adopted to evaluate attribute recognition performance \cite{li2018richly}. For the label-based metric, we compute the mean value of classification accuracy of positive samples and negative samples as the metric for each attribute. 
Then we take the average of all attributes as the mean accuracy (mA). The formulation is given as follows:
\begin{align}
    mA &= \frac{1}{M} \sum_{j=1}^M \frac{1}{2} (\frac{TP^{j}}{TP^{j} + FN^{j}} + \frac{TN^{j}}{TN^{j} + FP^{j}}) \nonumber \label{eq:ma}
\end{align}
For instance-based metrics, accuracy (Accu), precision (Prec), recall (Recall), and F1-score (F1) are used, which are computed as:
\begin{align}
    & Accu = \frac{1}{N} \sum_{i=1}^N \frac{TP_{i}}{TP_{i} + FP_{i} + FN_{i}}, \nonumber \\
    & Recall = \frac{1}{N} \sum_{i=1}^N \frac{TP_{i}}{TP_{i} + FN_{i}}, \nonumber \\
    & Prec = \frac{1}{N} \sum_{i=1}^N \frac{TP_{i}}{TP_{i} + FP_{i}}, \nonumber \\
    & F1 = \frac{1}{N} \sum_{i=1}^N \frac{2 \cdot Prec \cdot Recall}{Prec + Recall}, \nonumber
\end{align}
where $TP_{i}$, $FP_{i}$, $FN_{i}$ is the number of true positive, false positive, false negative attributes of $i$-th sample and $N$ denotes the number of samples.

\subsubsection{Experimental settings}
For pedestrian attribute recognition, our proposed method adopts ResNet-50 \cite{he2016deep} as the backbone network for a fair comparison. Pedestrian images are resized to $256 \times 192$ as inputs. Following the JLAC \cite{tan2020relation}, horizontal flipping, padding, random crop, random erasing, and random gaussian blurs are used as augmentations. SGD optimizer is employed for training with the weight decay of 0.0001. The initial learning rate equals 0.01, and the batch size is 64. Plateau learning rate scheduler with reduction factor 0.1 and loss patience 4 is adopted. The total epoch number of the training stage is 30.

\subsubsection{Performance comparison with SOTA methods}
We conduct experiments on $\text{PETA}_{\text{ZS}}$, $\text{RAP}_{\text{ZS}}$, and PA-100K datasets of pedestrian attribute recognition. The results are listed in Tab. \ref{table: pedes benchmark}. Compared with the baseline method, our method achieves 4.09, 6.07, and 3.97 performance improvement in the mA metric on $\text{PETA}_{\text{ZS}}$, $\text{RAP}_{\text{ZS}}$, and PA-100K datasets. Compared with previous method, we update SOTA performance and outperform the JLAC \cite{tan2020relation} by 1.92, 1.45, 2.47 on three datasets in the mA metric. The significant performance improvements on three datasets verify the superiority of our method.

\begin{figure}[t]
	\centering
	\subfloat[Angle distribution of positive samples in each attribute.]{
		\includegraphics[width=1\linewidth ]{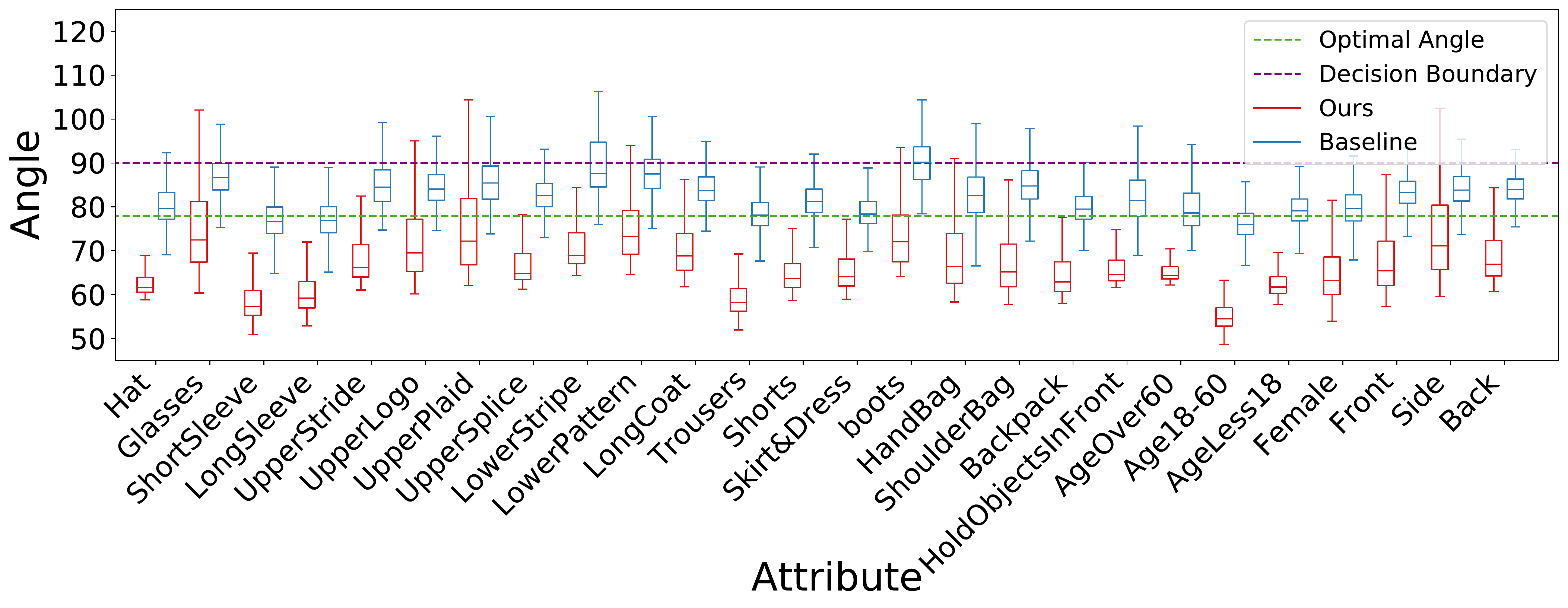}}  \\
	\subfloat[Angle distribution of negative samples in each attribute.]{
		\includegraphics[width=1\linewidth ]{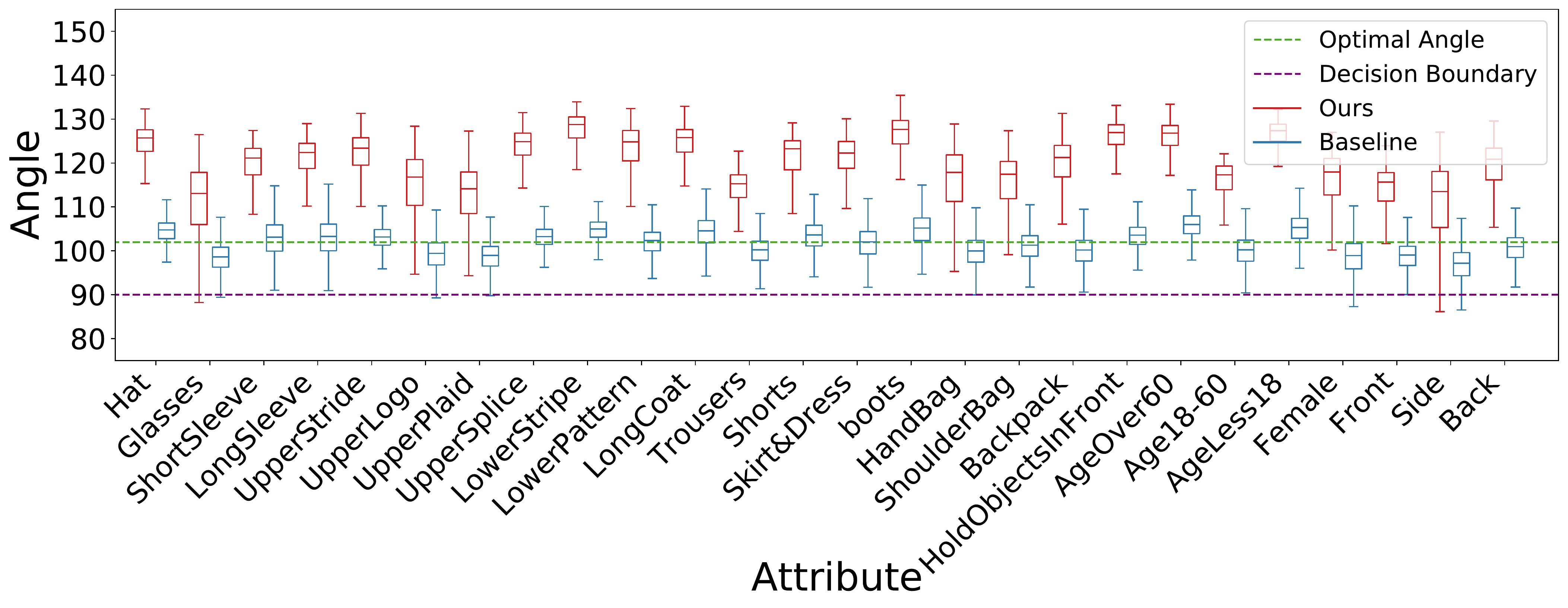}}
	\caption{\textbf{Angle distribution of the baseline model (blue box) and our method (red box) on PA-100K \cite{liu2017hydraplus}}. The green and purple dashed lines mark the angle decision boundary (90\degree) and the optimal angle of the OFML mechanism. We show angle distribution of positive samples (samples with corresponding attributes) in (a) and negative samples (samples without corresponding attributes) in (b). }
	\label{fig:angle_dist}
\end{figure}

\subsubsection{Visualization and Analysis}

Our DAFL framework aims to solve the problem of the OFML mechanism, that is, angles between features and classifiers are too large or even close to 90\degree. To visually demonstrate the progress made by our method, we take the results on PA-100K as an example and show the angles distribution of each attribute in Fig. \ref{fig:angle_dist}. We use ``positive (negative) samples'' of an attribute to denote the samples annotated with (without) corresponding attributes. On the one hand, compared to the baseline model, the angles of positive samples of our method DAFL are significantly smaller and farther away from the decision boundary (90\degree), as demonstrated in Fig. \ref{fig:angle_dist}(a). On the other hand, the angles of negative samples of our method are larger than the angles of the baseline model, as shown in Fig. \ref{fig:angle_dist}(b). The greater distance from the decision boundary in positive and negative samples means that our method yields higher confidence results and more robust predictions. In addition, most of the angles of the baseline model are distributed around two purple dashed lines, which is consistent with our theoretical analysis of the limitations of the OFML mechanism in Sec. \ref{sec:proof}.

Although our method has made remarkable progress, the angles between features and classifiers are still large. The reason is that the individual label features are not entirely disentangled from each other, \ie, they are not wholly independent. First, convolution kernels in the backbone network cause a feature overlap between adjacent features, which is inevitable in the image feature map $\mathcal{F} \in R^{C \times H \times W}$ in Fig. \ref{fig:framework}. Second, the label feature $\bm{F} \in R^{C \times M}$ is aggregated from the feature map $\mathcal{F}$. Due to the lack of precise semantic segmentation annotation, the aggregation relying on query attention map may incorporate the same pixel feature $\mathcal{F}_{h, w}$ into different label features. Therefore, we conclude that it is impossible to learn completely disentangled label features independent of each other. All we can do is decouple the label features as much as possible, so each label feature reflects its unique discriminative information.

\begin{figure}[t]
	\centering
	\subfloat[Ten images of the same pedestrian with a similar appearance.]{
		\includegraphics[width=1\linewidth ]{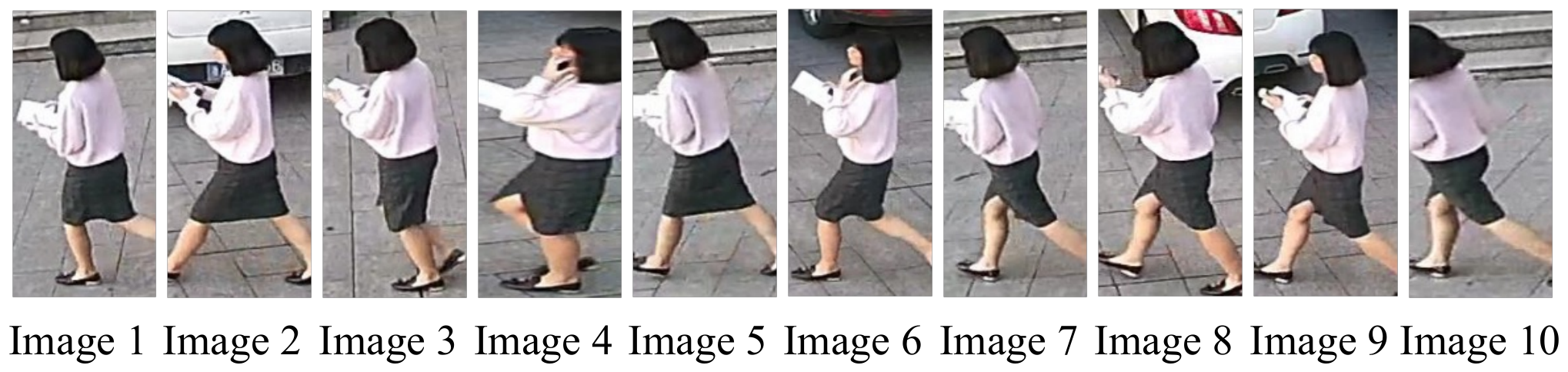}}  \\
	\subfloat[Predictions on ``Skirt\&Dress" and ``Female" attributes.]{
		\includegraphics[width=1\linewidth ]{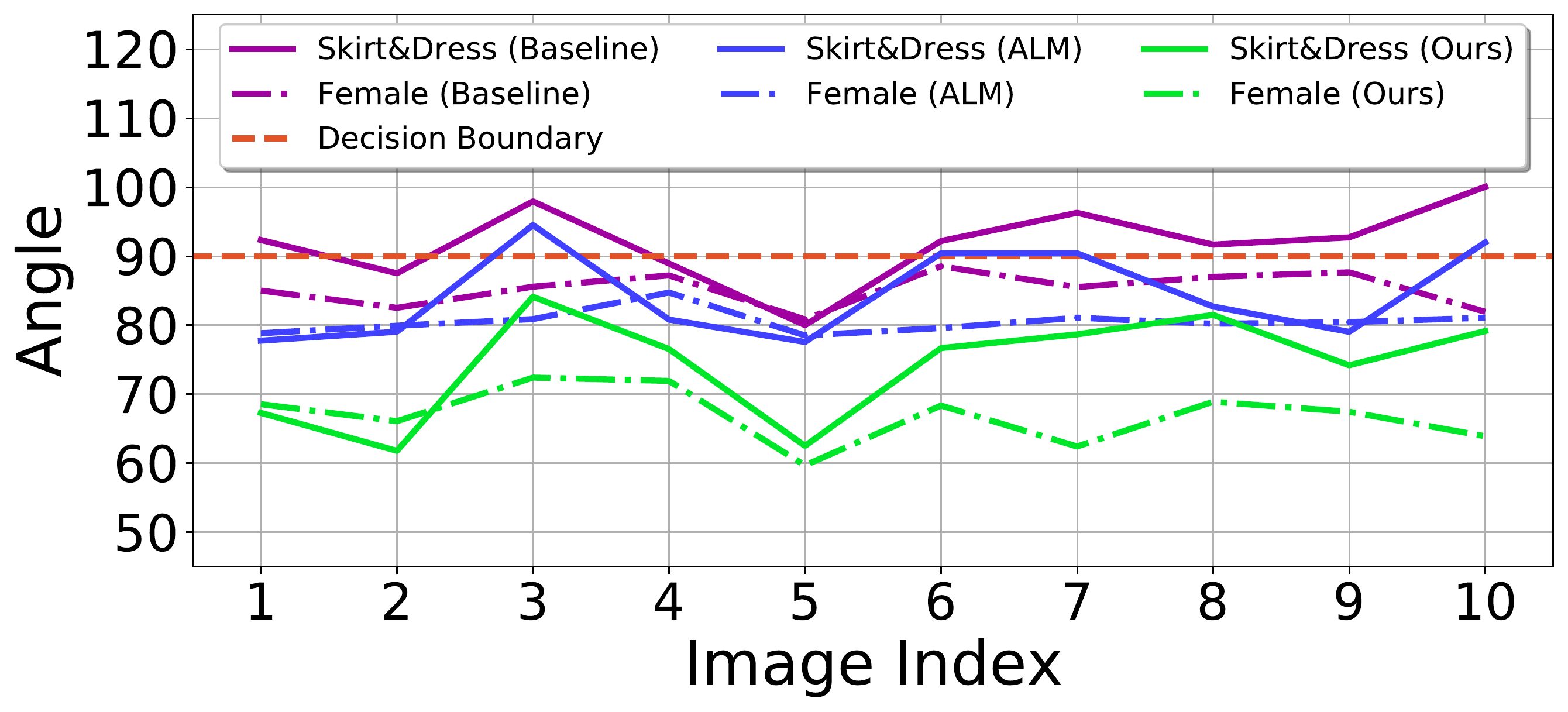}}
	\caption{\textbf{Predictions on images with marginal background and pose variation}. Due to the smaller angles between features and classifiers, our method shows stronger robustness than the baseline model.}
	\label{fig:cases}
\end{figure}

\begin{table*}[ht]
\centering{
\setlength\tabcolsep{4pt}
\caption{\textbf{Performance comparison with state-of-the-art methods on the $\text{COCO}_{\text{seq}}$}. The performance of our method is the average over five experiments.}
\label{table: COCOseq benchmark}
\renewcommand{\arraystretch}{1.2}
\begin{tabular}{c|ccc|ccc|ccc|ccc}
\toprule
\multirow{2}{*}{COCOseq} & \multicolumn{3}{c|}{majority} & \multicolumn{3}{c|}{moderate}& \multicolumn{3}{c|}{minority}& \multicolumn{3}{c}{Overall}\\ \cline{2-13}
 & C-F1 & O-F1 & mAP & C-F1 & O-F1 & mAP & C-F1 & O-F1 & mAP & C-F1 & O-F1 & mAP \\
\midrule
Multitask~\cite{caruana1997multitask} & 72.9 & 70.9 & 77.3 &  53.2  &  51.4 & 55.0 & 12.7 &  13.6 & 24.2 & 51.2 & 52.1  & 53.9 \\
Finetune  & 18.5 & 27.9 & 29.8 & 6.7  & 16.7 & 14.1 & 0.0  & 0.0  & 5.2  & 8.5  & 18.4  & 16.4 \\
EWC~\cite{kirkpatrick2017overcoming} & 60.0 & 53.4 & 64.1 & 37.3 & 38.1 & 47.5 &  7.5 & 8.2  & 21.5 & 38.9 & 40.0  & 46.6 \\
CRS~\cite{vitter1985random} & 67.0 & 62.5 & 67.9 & 47.8 & 45.2 & 50.4 & 14.5 & 15.6 & 26.9 & 47.5 & 46.6  & 50.2 \\
GSS~\cite{aljundi2019gradient} & 59.3 & 56.7 & 59.6 & 44.9 & 43.0 & 46.0 & 10.5 & 11.0 & 18.6 & 42.8 & 42.7 & 44.0 \\
ExStream~\cite{hayes2019memory} & 58.8 & 52.0 & 62.5 & 49.2 & 47.3 & 52.7 & 26.4 & 26.6 & 36.6 & 47.8 & 43.9 & 51.1 \\
PRS~\cite{kim2020imbalanced} & 65.4 & 59.3 & 67.5 & 52.5 & 49.7 & 55.2 & 34.5 & 34.6 & 39.7 & 53.2 & 50.3 & 55.3 \\ 
PRS* ~\cite{kim2020imbalanced} & 70.1 & 63.5 & 76.1 & 55.4 & 52.1 & 61.4 & 34.8 & 35.0 & 41.6 & 56.1 & 52.9 & 61.2 \\ \hline \hline
Ours + PRS & \bf{79.4} & \bf{77.2} & \bf{85.5} & \bf{69.3} & \bf{69.7} & \bf{74.6} & \bf{52.0} & \bf{52.8} & \bf{54.9} & \bf{68.7} & \bf{68.8} & \bf{73.4}  \\ \bottomrule
\end{tabular}
}
\end{table*}

\begin{table*}[ht]
\centering{
\setlength\tabcolsep{4pt}
\caption{\textbf{Performance comparison with state-of-the-art methods on the $\text{NUS-WIDE}_{\text{seq}}$}. The performance of our method is the average over five experiments.}
\label{table: NUSseq benchmark}
\renewcommand{\arraystretch}{1.2}
\begin{tabular}{c|ccc|ccc|ccc|ccc}
\toprule
\multirow{2}{*}{NUS-WIDEseq} & \multicolumn{3}{c|}{majority} & \multicolumn{3}{c|}{moderate}& \multicolumn{3}{c|}{minority} & \multicolumn{3}{c}{Overall}\\ \cline{2-13}
& C-F1 & O-F1 & mAP & C-F1 & O-F1 & mAP & C-F1 & O-F1 & mAP & C-F1 & O-F1 & mAP \\
\midrule
Multitask~\cite{caruana1997multitask} & 33.7 & 30.8 & 32.8 &  29.3  &  28.7 & 28.9 &  9.7 &  11.8 & 25.8 & 24.6 & 24.9  & 28.4 \\
Finetune  & 0.6 & 4.6 & 4.1 & 2.3 & 2.8 & 6.0 & 5.2 & 7.4 & 9.4 & 4.2 & 5.1 & 7.1 \\
EWC~\cite{kirkpatrick2017overcoming} & 15.7 & 9.9 & 15.8 & 16.3 & 12.6 & 19.4 & 12.3 & 13.9 & 24.1 & 17.1 & 11.4 & 20.7 \\
CRS~\cite{vitter1985random} & 28.4 & 17.8 & 21.9 & 13.6 & 14.2 & 18.5 & 10.4 & 11.8 & 20.6 & 16.8 & 15.0 & 20.1 \\
GSS~\cite{aljundi2019gradient} & 24.6 & 13.5 & 19.0 & 14.8 & 15.5 & 17.9 & 15.9 & 17.6 & 24.5 & 17.9 & 15.3 & 20.9 \\
ExStream~\cite{hayes2019memory} & 15.6 & 9.2 & 15.3 & 12.4 & 12.8 & 17.6 & 24.6 & 24.1 & 26.7 & 18.7 & 16.0 & 21.0 \\
PRS~\cite{kim2020imbalanced} & 26.7 & 17.9 & 21.2 & 19.2 & 19.3 & 21.5 & 27.5 & 26.8 & 31.0 & 24.8 & 21.7 & 25.5 \\
PRS* ~\cite{kim2020imbalanced} & 29.1 & 19.2 & 30.8 & 24.3 & 24.1 & 30.4 & 32.9 & 32.0 & 37.3 & 25.6 & 29.3 & 33.4 \\ \hline \hline
Ours + PRS & \bf{34.7} & \bf{32.5} & \bf{36.3} & \bf{37.5} & \bf{37.6} & \bf{41.5} & \bf{46.4} & \bf{46.2} & \bf{51.8} & \bf{40.1} & \bf{40.9} & \bf{44.9} \\ \bottomrule
\end{tabular}
}
\end{table*}

This property has significant implications in practical applications. Specific examples and corresponding prediction results are given in Fig. \ref{fig:cases}. For the ``Skirt\&Dress" attribute, our method predicts correct predictions in all ten images. Instead, the baseline model achieves correct predictions in only three images. The ALM \cite{tang2019Improving} model achieves correct predictions in six images. For the ``Female" attribute, although both methods achieve successful prediction on ten images, the prediction of our method is more confident with small angles. All these phenomena demonstrate the robustness of our method.

\subsection{Continual multi-label learning}

To further confirm the general applicability of our method, we apply our method to the continual multi-label learning task, a promising and valuable task solving the catastrophic forgetting problem in the context of multi-label classification. Taking sequential data steam of various tasks as input, continual learning accumulates knowledge over different tasks without the need to retrain from scratch \cite{delange2021continual}. For the general classification task, a batch of images is randomly sampled from all training data. Instead, continual multi-label learning first divides the training data into several tasks. Then, images of the current task are sampled to train the model until all images of the current task have been traversed. Tasks are traversed according to a predefined order. In this section, all experiments are conducted based on the protocol proposed in PRS \cite{kim2020imbalanced}.

\subsubsection{Datasets}
\textbf{$\text{COCO}_{seq}$} dataset \cite{kim2020imbalanced} is a 4-way split MSCOCO dataset \cite{lin2014microsoft} and consists of 35,072 training and 6,346 test images of 70 categories. For the training set, there are 8,365 images labeled with 24 categories, 7,814 with 18 categories, 11,603 with 14 categories, and 7,290 with 14 categories in task0, task1, task2, and task3, respectively. For the test set, four tasks have 1,960, 1,690, 1,400, and 1,316 images. Each category has 100 images in the test set. Since a single image contains multiple labels, the number of test sets is 6,346 not 7000. Following the definition \cite{kim2020imbalanced}, classes with less than 200 training examples are taken as the minority classes, 200-900 as moderate classes, and greater than 900 as the majority classes. These three types of classes are evaluated separately.

\textbf{$\text{NUS-WIDE}_{seq}$} dataset \cite{kim2020imbalanced} is a sequential dataset from NUS-WIDE \cite{chua2009nus}, containing 6 mutually exclusive and increasingly difficult tasks. $\text{NUS-WIDE}_{seq}$ dataset contains 48,724 training images and 2,367 test images with 49 categories. For the training set, 48,724 images are divided into 6 tasks, where tasks 0 to 5 has 17,672 (5), 18,402 (6), 3,623 (16), 1,948 (12), 2,373 (4), and 4,730 (6) images (categories), respectively. In the test set, there are 50 samples per category, for a total of 2,367 samples.

\subsubsection{Evaluation Metrics}
To make a fair comparison, we report the three most important performance metrics in the generic multi-label classification task, \ie, mean average precision (mAP), per-class F1 (CF1), and overall F1 (OF1). Besides the performance on all classes like the generic multi-label classification task, the performance of continual multi-label learning is also evaluated in three parts: majority classes, moderate classes, and minority classes.

\subsubsection{Experimental settings}

For the continual multi-label learning, we follow the same experimental setting as PRS \cite{kim2020imbalanced}. We insert our SSCA module between the backbone and classifier of the PRS framework and keep other components unchanged. We also adopt ResNet-101 \cite{he2016deep} as the backbone network. Image on \textbf{$\text{COCO}_{seq}$} and \textbf{$\text{NUS-WIDE}_{seq}$} are resized to $256 \times 256$ as inputs. SGD optimizer is employed for training with the weight decay of 0.0001. The initial learning rate equals 0.0005. Details of hyper-parameters setting can be found in \cite{kim2020imbalanced}. To avoid random noise, all performances on the continuous multi-label learning task are the mean value of five experiments.

\subsubsection{Performance comparison with SOTA methods}

In this section, experiments are conducted following the exact same settings proposed in PRS \cite{kim2020imbalanced}. To make a fair comparison, we reproduce the PRS results as PRS* in Tab. \ref{table: COCOseq benchmark} and Tab. \ref{table: NUSseq benchmark}. As shown in Tab. \ref{table: COCOseq benchmark} and Tab. \ref{table: NUSseq benchmark}, our method achieves a much higher performance on all metrics of two datasets. The performance improvements on both datasets validate the effectiveness and applicability of our approach as a plug-and-play module. To avoid random noise, the reported performance of our method is the average over five experimental runs.

\section{Conclusion and future work} \label{sec:conclusion}
This work reveals the limitations of the One-shared-Feature-for-Multiple-Labels (OFML) mechanism adopted by most existing works. The OFML mechanism adopts a learned image feature to classify multiple labels, ignoring the distinction between multi-label and single-label (multi-class) classification. We mathematically prove that the image feature learned by the OFML mechanism cannot maintain high similarity with multiple classifier weights simultaneously, making the model non-robust and getting low confidence predictions. To solve the limitations, we introduce the One-specific-Feature-for-One-Label (OFOL) mechanism and propose the Semantic Spatial Cross-Attention (SSCA) module to extract a specific discriminative feature for each label. The SSCA module locates label-related regions based on learnable semantic queries and aggregates located region features into the corresponding label feature. We conduct exhaustive experiments on the eight datasets of generic multi-label classification, pedestrian attribute recognition, and continual multi-label learning tasks. The state-of-the-art performance on generic multi-label classification and pedestrian attribute recognition show the superiority of our method. The performance improvement on the continual multi-label learning task validates the applicability of our method as a plug-and-play module on multi-label related tasks.

It is worth noting that our SSCA module is only one instance of the feature disentangling module in the Disentangled Label Feature Learning (DLFL) framework. To validate the effectiveness of the OFOL mechanism, our SSCA module is straightforward and introduces as few parameters and calculations as possible. We believe that, following the OFOL mechanism, researchers can design powerful instances of the feature disentangling module and make a further step to disentangle label features and improve feature representation.

\section{ACKNOWLEDGMENTS} 

This work is supported in part by the National Natural Science Foundation of China (Grant No. 61721004 and No.62176255), the Projects of Chinese Academy of Science (Grant No. QYZDB-SSW-JSC006), the Strategic Priority Research Program of Chinese Academy of Sciences (Grant No. XDA27000000) and the Youth Innovation Promotion Association CAS.

\ifCLASSOPTIONcaptionsoff
  \newpage
\fi


\bibliographystyle{IEEEtran}
\bibliography{egbib}

\begin{thebibliography}{10}
\providecommand{\url}[1]{#1}
\csname url@samestyle\endcsname
\providecommand{\newblock}{\relax}
\providecommand{\bibinfo}[2]{#2}
\providecommand{\BIBentrySTDinterwordspacing}{\spaceskip=0pt\relax}
\providecommand{\BIBentryALTinterwordstretchfactor}{4}
\providecommand{\BIBentryALTinterwordspacing}{\spaceskip=\fontdimen2\font plus
\BIBentryALTinterwordstretchfactor\fontdimen3\font minus
  \fontdimen4\font\relax}
\providecommand{\BIBforeignlanguage}[2]{{%
\expandafter\ifx\csname l@#1\endcsname\relax
\typeout{** WARNING: IEEEtran.bst: No hyphenation pattern has been}%
\typeout{** loaded for the language `#1'. Using the pattern for}%
\typeout{** the default language instead.}%
\else
\language=\csname l@#1\endcsname
\fi
#2}}
\providecommand{\BIBdecl}{\relax}
\BIBdecl

\bibitem{read2011classifier}
J.~Read, B.~Pfahringer, G.~Holmes, and E.~Frank, ``Classifier chains for
  multi-label classification,'' \emph{Machine learning}, vol.~85, no.~3, pp.
  333--359, 2011.

\bibitem{zhang2013review}
M.-L. Zhang and Z.-H. Zhou, ``A review on multi-label learning algorithms,''
  \emph{IEEE transactions on knowledge and data engineering}, vol.~26, no.~8,
  pp. 1819--1837, 2013.

\bibitem{yang2016exploit}
H.~Yang, J.~Tianyi~Zhou, Y.~Zhang, B.-B. Gao, J.~Wu, and J.~Cai, ``Exploit
  bounding box annotations for multi-label object recognition,'' in
  \emph{Proceedings of the IEEE Conference on Computer Vision and Pattern
  Recognition}, 2016, pp. 280--288.

\bibitem{liu2021emerging}
W.~Liu, H.~Wang, X.~Shen, and I.~Tsang, ``The emerging trends of multi-label
  learning,'' \emph{IEEE transactions on pattern analysis and machine
  intelligence}, 2021.

\bibitem{lo2011cost}
H.-Y. Lo, J.-C. Wang, H.-M. Wang, and S.-D. Lin, ``Cost-sensitive multi-label
  learning for audio tag annotation and retrieval,'' \emph{IEEE Transactions on
  Multimedia}, vol.~13, no.~3, pp. 518--529, 2011.

\bibitem{li2018richly}
D.~Li, Z.~Zhang, X.~Chen, and K.~Huang, ``A richly annotated pedestrian dataset
  for person retrieval in real surveillance scenarios,'' \emph{IEEE Trans.
  Image Process.}, vol.~28, no.~4, pp. 1575--1590, 2018.

\bibitem{chen2021multi}
B.~Chen, Z.~Zhang, Y.~Li, G.~Lu, and D.~Zhang, ``Multi-label chest x-ray image
  classification via semantic similarity graph embedding,'' \emph{IEEE
  Transactions on Circuits and Systems for Video Technology}, 2021.

\bibitem{Wang_2020_CVPR}
D.~Wang and S.~Zhang, ``Unsupervised person re-identification via multi-label
  classification,'' in \emph{Proceedings of the IEEE/CVF Conference on Computer
  Vision and Pattern Recognition (CVPR)}, June 2020.

\bibitem{ridnik2021asymmetric}
T.~Ridnik, E.~Ben-Baruch, N.~Zamir, A.~Noy, I.~Friedman, M.~Protter, and
  L.~Zelnik-Manor, ``Asymmetric loss for multi-label classification,'' in
  \emph{Proceedings of the IEEE/CVF International Conference on Computer
  Vision}, 2021, pp. 82--91.

\bibitem{chen2019multi}
Z.-M. Chen, X.-S. Wei, P.~Wang, and Y.~Guo, ``Multi-label image recognition
  with graph convolutional networks,'' in \emph{Proc. IEEE Conf. Comput. Vis.
  Pattern Recognit.}, 2019, pp. 5177--5186.

\bibitem{chen2021P_GCN}
Z.~Chen, X.-S. Wei, P.~Wang, and Y.~Guo, ``Learning graph convolutional
  networks for multi-label recognition and applications,'' \emph{IEEE
  Transactions on Pattern Analysis and Machine Intelligence}, 2021.

\bibitem{liu2017hydraplus}
X.~Liu, H.~Zhao, M.~Tian, L.~Sheng, J.~Shao, S.~Yi, J.~Yan, and X.~Wang,
  ``Hydraplus-net: Attentive deep features for pedestrian analysis,'' in
  \emph{Proc. IEEE Int. Conf. Comput. Vis.}, 2017, pp. 350--359.

\bibitem{tang2019Improving}
C.~Tang, L.~Sheng, Z.~Zhang, and X.~Hu, ``Improving pedestrian attribute
  recognition with weakly-supervised multi-scale attribute-speciﬁc
  localization,'' in \emph{Proc. IEEE Int. Conf. Comput. Vis.}, 2019.

\bibitem{sarafianos2018deep}
N.~Sarafianos, X.~Xu, and I.~A. Kakadiaris, ``Deep imbalanced attribute
  classification using visual attention aggregation,'' in \emph{Proc. Eur.
  Conf. Comput. Vis.}, 2018, pp. 680--697.

\bibitem{guo2019visual}
H.~Guo, K.~Zheng, X.~Fan, H.~Yu, and S.~Wang, ``Visual attention consistency
  under image transforms for multi-label image classification,'' in \emph{Proc.
  IEEE Conf. Comput. Vis. Pattern Recognit.}, 2019, pp. 729--739.

\bibitem{chen2022sst}
Z.-M. Chen, Q.~Cui, B.~Zhao, R.~Song, X.~Zhang, and O.~Yoshie, ``Sst: Spatial
  and semantic transformers for multi-label image recognition,'' \emph{IEEE
  Transactions on Image Processing}, vol.~31, pp. 2570--2583, 2022.

\bibitem{gao2021MCAR}
B.-B. Gao and H.-Y. Zhou, ``Learning to discover multi-class attentional
  regions for multi-label image recognition,'' \emph{IEEE Transactions on Image
  Processing}, vol.~30, pp. 5920--5932, 2021.

\bibitem{zhu2021residual}
K.~Zhu and J.~Wu, ``Residual attention: A simple but effective method for
  multi-label recognition,'' in \emph{Proceedings of the IEEE/CVF International
  Conference on Computer Vision}, 2021, pp. 184--193.

\bibitem{zhu2017learning}
F.~Zhu, H.~Li, W.~Ouyang, N.~Yu, and X.~Wang, ``Learning spatial regularization
  with image-level supervisions for multi-label image classification,'' in
  \emph{Proceedings of the IEEE Conference on Computer Vision and Pattern
  Recognition}, 2017, pp. 5513--5522.

\bibitem{deng2009imagenet}
J.~Deng, W.~Dong, R.~Socher, L.-J. Li, K.~Li, and L.~Fei-Fei, ``Imagenet: A
  large-scale hierarchical image database,'' in \emph{2009 IEEE conference on
  computer vision and pattern recognition}.\hskip 1em plus 0.5em minus
  0.4em\relax Ieee, 2009, pp. 248--255.

\bibitem{simonyan2014very}
K.~Simonyan and A.~Zisserman, ``Very deep convolutional networks for
  large-scale image recognition,'' \emph{arXiv preprint arXiv:1409.1556}, 2014.

\bibitem{he2016deep}
K.~He, X.~Zhang, S.~Ren, and J.~Sun, ``Deep residual learning for image
  recognition,'' in \emph{Proc. IEEE Conf. Comput. Vis. Pattern Recognit.},
  2016, pp. 770--778.

\bibitem{huang2017densely}
G.~Huang, Z.~Liu, L.~Van Der~Maaten, and K.~Q. Weinberger, ``Densely connected
  convolutional networks,'' in \emph{Proceedings of the IEEE conference on
  computer vision and pattern recognition}, 2017, pp. 4700--4708.

\bibitem{hu2018squeeze}
J.~Hu, L.~Shen, and G.~Sun, ``Squeeze-and-excitation networks,'' in
  \emph{Proceedings of the IEEE conference on computer vision and pattern
  recognition}, 2018, pp. 7132--7141.

\bibitem{lin2014microsoft}
T.-Y. Lin, M.~Maire, S.~Belongie, J.~Hays, P.~Perona, D.~Ramanan,
  P.~Doll{\'a}r, and C.~L. Zitnick, ``Microsoft coco: Common objects in
  context,'' in \emph{Proc. Eur. Conf. Comput. Vis.}\hskip 1em plus 0.5em minus
  0.4em\relax Springer, 2014, pp. 740--755.

\bibitem{everingham2010pascal}
M.~Everingham, L.~Van~Gool, C.~K. Williams, J.~Winn, and A.~Zisserman, ``The
  pascal visual object classes (voc) challenge,'' \emph{Int. J. Comput. Vis},
  vol.~88, no.~2, pp. 303--338, 2010.

\bibitem{chua2009nus}
T.-S. Chua, J.~Tang, R.~Hong, H.~Li, Z.~Luo, and Y.~Zheng, ``Nus-wide: a
  real-world web image database from national university of singapore,'' in
  \emph{Proceedings of the ACM international conference on image and video
  retrieval}, 2009, pp. 1--9.

\bibitem{jia2021rethinking}
J.~Jia, H.~Huang, X.~Chen, and K.~Huang, ``Rethinking of pedestrian attribute
  recognition: A reliable evaluation under zero-shot pedestrian identity
  setting,'' \emph{arXiv preprint arXiv:2107.03576}, 2021.

\bibitem{kim2020imbalanced}
C.~D. Kim, J.~Jeong, and G.~Kim, ``Imbalanced continual learning with
  partitioning reservoir sampling,'' in \emph{European Conference on Computer
  Vision}.\hskip 1em plus 0.5em minus 0.4em\relax Springer, 2020, pp. 411--428.

\bibitem{jia2022learning}
J.~Jia, N.~Gao, F.~He, X.~Chen, and K.~Huang, ``Learning disentangled attribute
  representations for robust pedestrian attribute recognition,'' in \emph{Proc.
  AAAI Conf. Artif. Intell.}, 2022.

\bibitem{wei2015hcp}
Y.~Wei, W.~Xia, M.~Lin, J.~Huang, B.~Ni, J.~Dong, Y.~Zhao, and S.~Yan, ``Hcp: A
  flexible cnn framework for multi-label image classification,'' \emph{IEEE
  transactions on pattern analysis and machine intelligence}, vol.~38, no.~9,
  pp. 1901--1907, 2015.

\bibitem{wang2016beyond}
M.~Wang, C.~Luo, R.~Hong, J.~Tang, and J.~Feng, ``Beyond object proposals:
  Random crop pooling for multi-label image recognition,'' \emph{IEEE
  Transactions on Image Processing}, vol.~25, no.~12, pp. 5678--5688, 2016.

\bibitem{wang2017multi}
Z.~Wang, T.~Chen, G.~Li, R.~Xu, and L.~Lin, ``Multi-label image recognition by
  recurrently discovering attentional regions,'' in \emph{Proceedings of the
  IEEE international conference on computer vision}, 2017, pp. 464--472.

\bibitem{jaderberg2015spatial}
M.~Jaderberg, K.~Simonyan, A.~Zisserman \emph{et~al.}, ``Spatial transformer
  networks,'' in \emph{Proc. Adv. Neural Inf. Process. Syst.}, 2015, pp.
  2017--2025.

\bibitem{vaswani2017attention}
A.~Vaswani, N.~Shazeer, N.~Parmar, J.~Uszkoreit, L.~Jones, A.~N. Gomez, L.~u.
  Kaiser, and I.~Polosukhin, ``Attention is all you need,'' in \emph{Advances
  in Neural Information Processing Systems}, vol.~30, 2017.

\bibitem{hochreiter1997long}
S.~Hochreiter and J.~Schmidhuber, ``Long short-term memory,'' \emph{Neural
  computation}, vol.~9, no.~8, pp. 1735--1780, 1997.

\bibitem{chen2019learning}
T.~Chen, M.~Xu, X.~Hui, H.~Wu, and L.~Lin, ``Learning semantic-specific graph
  representation for multi-label image recognition,'' in \emph{Proceedings of
  the IEEE/CVF International Conference on Computer Vision}, 2019, pp.
  522--531.

\bibitem{li2015gated}
Y.~Li, D.~Tarlow, M.~Brockschmidt, and R.~Zemel, ``Gated graph sequence neural
  networks,'' \emph{arXiv preprint arXiv:1511.05493}, 2015.

\bibitem{kipf2017semi}
T.~N. Kipf and M.~Welling, ``Semi-supervised classification with graph
  convolutional networks,'' in \emph{Int. Conf. Learn. Representations.}, 2017.

\bibitem{ye2020attention}
J.~Ye, J.~He, X.~Peng, W.~Wu, and Y.~Qiao, ``Attention-driven dynamic graph
  convolutional network for multi-label image recognition,'' in \emph{European
  Conference on Computer Vision}.\hskip 1em plus 0.5em minus 0.4em\relax
  Springer, 2020, pp. 649--665.

\bibitem{zhao2021transformer}
J.~Zhao, K.~Yan, Y.~Zhao, X.~Guo, F.~Huang, and J.~Li, ``Transformer-based dual
  relation graph for multi-label image recognition,'' in \emph{Proceedings of
  the IEEE/CVF International Conference on Computer Vision}, 2021, pp.
  163--172.

\bibitem{lanchantin2021general}
J.~Lanchantin, T.~Wang, V.~Ordonez, and Y.~Qi, ``General multi-label image
  classification with transformers,'' in \emph{Proceedings of the IEEE/CVF
  Conference on Computer Vision and Pattern Recognition}, 2021, pp.
  16\,478--16\,488.

\bibitem{li2015deepmar}
D.~Li, X.~Chen, and K.~Huang, ``Multi-attribute learning for pedestrian
  attribute recognition in surveillance scenarios,'' in \emph{Proc. IEEE Asia
  Conf. Pattern Recognit.}, 2015, pp. 111--115.

\bibitem{yu2016weakly}
K.~Yu, B.~Leng, Z.~Zhang, D.~Li, and K.~Huang, ``Weakly-supervised learning of
  mid-level features for pedestrian attribute recognition and localization,''
  in \emph{Proc. Brit. Mach. Vis. Conf}, 2017.

\bibitem{li2018pose}
D.~Li, X.~Chen, Z.~Zhang, and K.~Huang, ``Pose guided deep model for pedestrian
  attribute recognition in surveillance scenarios,'' in \emph{Proc. IEEE Int.
  Conf. Multimedia Expo.}\hskip 1em plus 0.5em minus 0.4em\relax IEEE, 2018,
  pp. 1--6.

\bibitem{liu2018localization}
P.~Liu, X.~Liu, J.~Yan, and J.~Shao, ``Localization guided learning for
  pedestrian attribute recognition,'' in \emph{Proc. Brit. Mach. Vis. Conf},
  2018.

\bibitem{han2019attribute}
K.~Han, Y.~Wang, H.~Shu, C.~Liu, C.~Xu, and C.~Xu, ``Attribute aware pooling
  for pedestrian attribute recognition,'' in \emph{Proc. Int. Joint Conf.
  Artif. Intell.}, 2019.

\bibitem{tan2020relation}
Z.~Tan, Y.~Yang, J.~Wan, G.~Guo, and S.~Z. Li, ``Relation-aware pedestrian
  attribute recognition with graph convolutional networks,'' in \emph{Proc.
  AAAI Conf. Artif. Intell.}, vol.~34, no.~07, 2020, pp. 12\,055--12\,062.

\bibitem{jia2021spatial}
J.~Jia, X.~Chen, and K.~Huang, ``Spatial and semantic consistency
  regularizations for pedestrian attribute recognition,'' in \emph{Proceedings
  of the IEEE/CVF International Conference on Computer Vision}, 2021, pp.
  962--971.

\bibitem{zhao2018grouping}
X.~Zhao, L.~Sang, G.~Ding, Y.~Guo, and X.~Jin, ``Grouping attribute recognition
  for pedestrian with joint recurrent learning.'' in \emph{Proc. Int. Joint
  Conf. Artif. Intell.}, 2018, pp. 3177--3183.

\bibitem{li2019pedestrian}
Q.~Li, X.~Zhao, R.~He, and K.~Huang, ``Pedestrian attribute recognition by
  joint visual-semantic reasoning and knowledge distillation.'' in \emph{Proc.
  Int. Joint Conf. Artif. Intell.}, 2019, pp. 833--839.

\bibitem{wang2017attribute}
J.~Wang, X.~Zhu, S.~Gong, and W.~Li, ``Attribute recognition by joint recurrent
  learning of context and correlation,'' in \emph{Proc. IEEE Int. Conf. Comput.
  Vis.}, 2017, pp. 531--540.

\bibitem{li2019visual}
Q.~Li, X.~Zhao, R.~He, and K.~Huang, ``Visual-semantic graph reasoning for
  pedestrian attribute recognition,'' in \emph{Proc. AAAI Conf. Artif.
  Intell.}, vol.~33, no.~01, 2019, pp. 8634--8641.

\bibitem{lin2017feature}
T.-Y. Lin, P.~Doll{\'a}r, R.~Girshick, K.~He, B.~Hariharan, and S.~Belongie,
  ``Feature pyramid networks for object detection,'' in \emph{Proc. IEEE Conf.
  Comput. Vis. Pattern Recognit.}, 2017, pp. 2117--2125.

\bibitem{li2017learning}
Z.~Li and D.~Hoiem, ``Learning without forgetting,'' \emph{IEEE transactions on
  pattern analysis and machine intelligence}, vol.~40, no.~12, pp. 2935--2947,
  2017.

\bibitem{yang2022uncertainty}
G.~Yang, E.~Fini, D.~Xu, P.~Rota, M.~Ding, M.~Nabi, X.~Alameda-Pineda, and
  E.~Ricci, ``Uncertainty-aware contrastive distillation for incremental
  semantic segmentation,'' \emph{IEEE Transactions on Pattern Analysis and
  Machine Intelligence}, 2022.

\bibitem{zhao2021mgsvf}
H.~Zhao, Y.~Fu, M.~Kang, Q.~Tian, F.~Wu, and X.~Li, ``Mgsvf: Multi-grained slow
  vs. fast framework for few-shot class-incremental learning,'' \emph{IEEE
  Transactions on Pattern Analysis and Machine Intelligence}, 2021.

\bibitem{kj2021incremental}
J.~Kj, J.~Rajasegaran, S.~Khan, F.~S. Khan, and V.~N. Balasubramanian,
  ``Incremental object detection via meta-learning,'' \emph{IEEE Transactions
  on Pattern Analysis and Machine Intelligence}, 2021.

\bibitem{park2021class}
J.~Park, M.~Kang, and B.~Han, ``Class-incremental learning for action
  recognition in videos,'' in \emph{Proceedings of the IEEE/CVF International
  Conference on Computer Vision}, 2021, pp. 13\,698--13\,707.

\bibitem{vershynin2018high}
R.~Vershynin, \emph{High-dimensional probability: An introduction with
  applications in data science}.\hskip 1em plus 0.5em minus 0.4em\relax
  Cambridge university press, 2018, vol.~47.

\bibitem{he2015delving}
K.~He, X.~Zhang, S.~Ren, and J.~Sun, ``Delving deep into rectifiers: Surpassing
  human-level performance on imagenet classification,'' in \emph{Proceedings of
  the IEEE international conference on computer vision}, 2015, pp. 1026--1034.

\bibitem{song2018deep}
L.~Song, J.~Liu, B.~Qian, M.~Sun, K.~Yang, M.~Sun, and S.~Abbas, ``A deep
  multi-modal cnn for multi-instance multi-label image classification,''
  \emph{IEEE Transactions on Image Processing}, vol.~27, no.~12, pp.
  6025--6038, 2018.

\bibitem{liu2021swin}
Z.~Liu, Y.~Lin, Y.~Cao, H.~Hu, Y.~Wei, Z.~Zhang, S.~Lin, and B.~Guo, ``Swin
  transformer: Hierarchical vision transformer using shifted windows,'' in
  \emph{Proceedings of the IEEE/CVF International Conference on Computer
  Vision}, 2021, pp. 10\,012--10\,022.

\bibitem{wang2016cnn}
J.~Wang, Y.~Yang, J.~Mao, Z.~Huang, C.~Huang, and W.~Xu, ``Cnn-rnn: A unified
  framework for multi-label image classification,'' in \emph{Proceedings of the
  IEEE conference on computer vision and pattern recognition}, 2016, pp.
  2285--2294.

\bibitem{chen2018order}
S.-F. Chen, Y.-C. Chen, C.-K. Yeh, and Y.-C. Wang, ``Order-free rnn with visual
  attention for multi-label classification,'' in \emph{Proceedings of the AAAI
  Conference on Artificial Intelligence}, vol.~32, no.~1, 2018.

\bibitem{yazici2020orderless}
V.~O. Yazici, A.~Gonzalez-Garcia, A.~Ramisa, B.~Twardowski, and J.~v.~d.
  Weijer, ``Orderless recurrent models for multi-label classification,'' in
  \emph{Proceedings of the IEEE/CVF Conference on Computer Vision and Pattern
  Recognition}, 2020, pp. 13\,440--13\,449.

\bibitem{wang2020multi}
Y.~Wang, D.~He, F.~Li, X.~Long, Z.~Zhou, J.~Ma, and S.~Wen, ``Multi-label
  classification with label graph superimposing,'' in \emph{Proceedings of the
  AAAI Conference on Artificial Intelligence}, vol.~34, no.~07, 2020, pp.
  12\,265--12\,272.

\bibitem{chen2021learning}
Z.~Chen, X.-S. Wei, P.~Wang, and Y.~Guo, ``Learning graph convolutional
  networks for multi-label recognition and applications,'' \emph{IEEE
  Transactions on Pattern Analysis and Machine Intelligence}, 2021.

\bibitem{lanchantin2021CTran}
J.~Lanchantin, T.~Wang, V.~Ordonez, and Y.~Qi, ``General multi-label image
  classification with transformers,'' in \emph{Proceedings of the IEEE/CVF
  Conference on Computer Vision and Pattern Recognition}, 2021, pp.
  16\,478--16\,488.

\bibitem{chen2018recurrent}
T.~Chen, Z.~Wang, G.~Li, and L.~Lin, ``Recurrent attentional reinforcement
  learning for multi-label image recognition,'' in \emph{Proceedings of the
  AAAI Conference on Artificial Intelligence}, vol.~32, no.~1, 2018.

\bibitem{gong2013deep}
Y.~Gong, Y.~Jia, T.~Leung, A.~Toshev, and S.~Ioffe, ``Deep convolutional
  ranking for multilabel image annotation,'' \emph{arXiv preprint
  arXiv:1312.4894}, 2013.

\bibitem{lee2018multi}
C.-W. Lee, W.~Fang, C.-K. Yeh, and Y.-C.~F. Wang, ``Multi-label zero-shot
  learning with structured knowledge graphs,'' in \emph{Proceedings of the IEEE
  conference on computer vision and pattern recognition}, 2018, pp. 1576--1585.

\bibitem{deng2014pedestrian}
Y.~Deng, P.~Luo, C.~C. Loy, and X.~Tang, ``Pedestrian attribute recognition at
  far distance,'' in \emph{Proc. ACM Int. Conf. Multimedia}.\hskip 1em plus
  0.5em minus 0.4em\relax ACM, 2014, pp. 789--792.

\bibitem{caruana1997multitask}
R.~Caruana, ``Multitask learning,'' \emph{Machine learning}, vol.~28, no.~1,
  pp. 41--75, 1997.

\bibitem{kirkpatrick2017overcoming}
J.~Kirkpatrick, R.~Pascanu, N.~Rabinowitz, J.~Veness, G.~Desjardins, A.~A.
  Rusu, K.~Milan, J.~Quan, T.~Ramalho, A.~Grabska-Barwinska \emph{et~al.},
  ``Overcoming catastrophic forgetting in neural networks,'' \emph{Proceedings
  of the national academy of sciences}, vol. 114, no.~13, pp. 3521--3526, 2017.

\bibitem{vitter1985random}
J.~S. Vitter, ``Random sampling with a reservoir,'' \emph{ACM Transactions on
  Mathematical Software (TOMS)}, vol.~11, no.~1, pp. 37--57, 1985.

\bibitem{aljundi2019gradient}
R.~Aljundi, M.~Lin, B.~Goujaud, and Y.~Bengio, ``Gradient based sample
  selection for online continual learning,'' \emph{Advances in neural
  information processing systems}, vol.~32, 2019.

\bibitem{hayes2019memory}
T.~L. Hayes, N.~D. Cahill, and C.~Kanan, ``Memory efficient experience replay
  for streaming learning,'' in \emph{2019 International Conference on Robotics
  and Automation (ICRA)}.\hskip 1em plus 0.5em minus 0.4em\relax IEEE, 2019,
  pp. 9769--9776.

\bibitem{delange2021continual}
M.~Delange, R.~Aljundi, M.~Masana, S.~Parisot, X.~Jia, A.~Leonardis,
  G.~Slabaugh, and T.~Tuytelaars, ``A continual learning survey: Defying
  forgetting in classification tasks,'' \emph{IEEE Transactions on Pattern
  Analysis and Machine Intelligence}, 2021.

\end{thebibliography}

%



%

\begin{IEEEbiography}[{\includegraphics[width=1in,height=1.25in,clip,keepaspectratio]{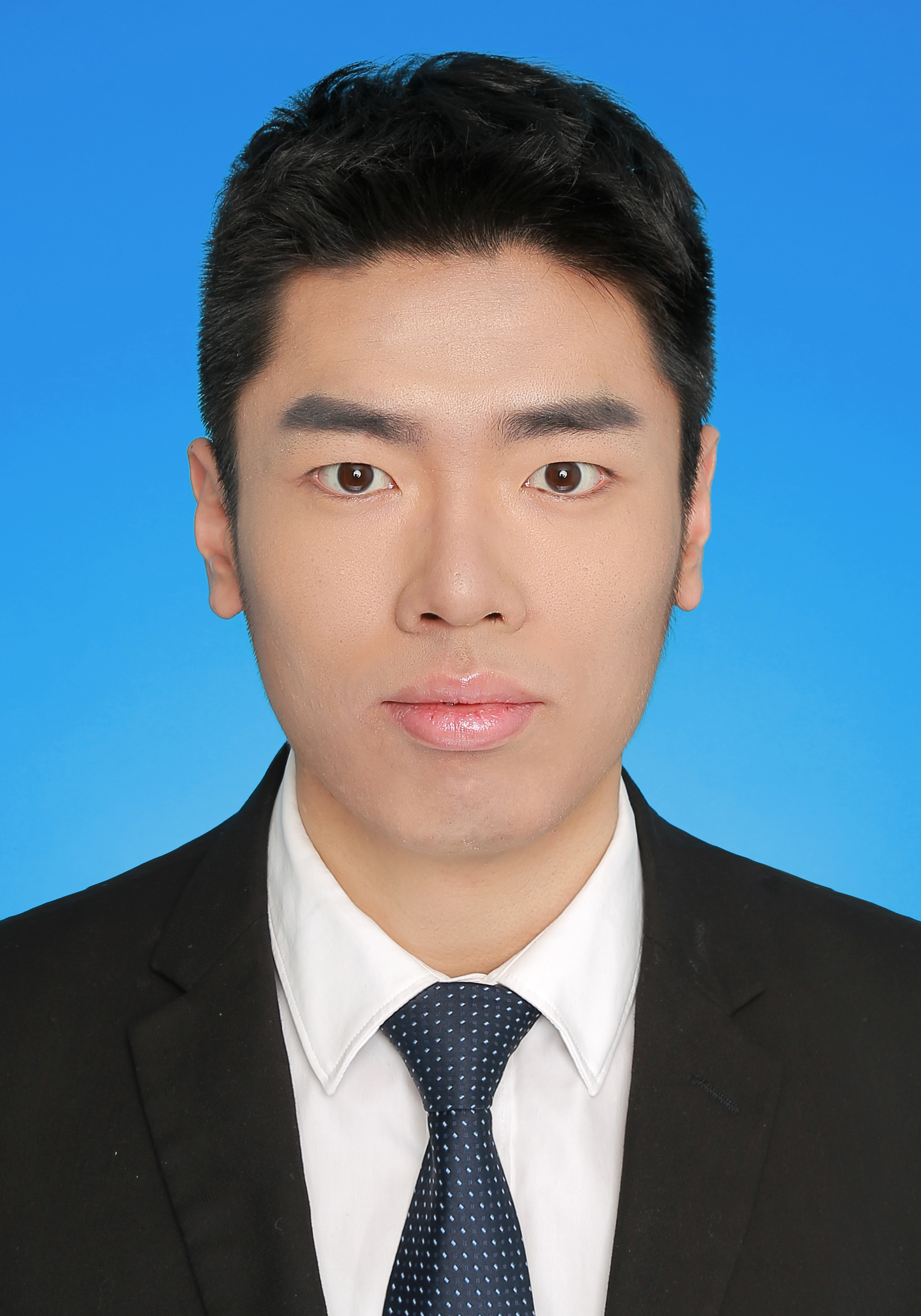}}]{Jian Jia} received the B.E. degree from Shandong University (SDU) in 2015 and the M.E. degree from Beijing University of Posts and Telecommunications (BUPT) in 2018. He is currently pursuing the Ph.D. degree with the School of Artificial Intelligence, University of Chinese Academy of Sciences(UCAS), Beijing 100049, China, and also the Center for Research on Intelligent System and Engineering (CRISE),Institute of Automation, Chinese Academy of Sciences (CASIA), Beijing 100190, China. His research interests include computer vision, deep learning, pedestrian attribute recognition and multi-label classification. 
\end{IEEEbiography}

\begin{IEEEbiography}[{\includegraphics[width=1in,height=1.25in,clip,keepaspectratio]{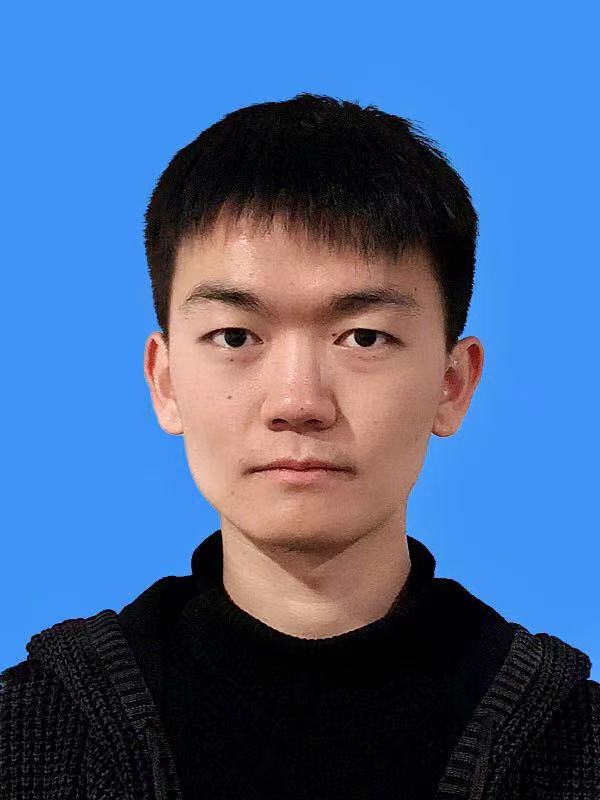}}]{Fei He} received the B.E. degree in the department of automation from University of Science and Technology of China (USTC), Hefei, China, in 2018. He is currently pursuing the Ph.D. degree with the Institute of Automation, Chinese Academy of Sciences (CASIA), Beijing, China. His research interests include computer vision and deep learning.
\end{IEEEbiography}

\begin{IEEEbiography}[{\includegraphics[width=1in,height=1.25in,clip,keepaspectratio]{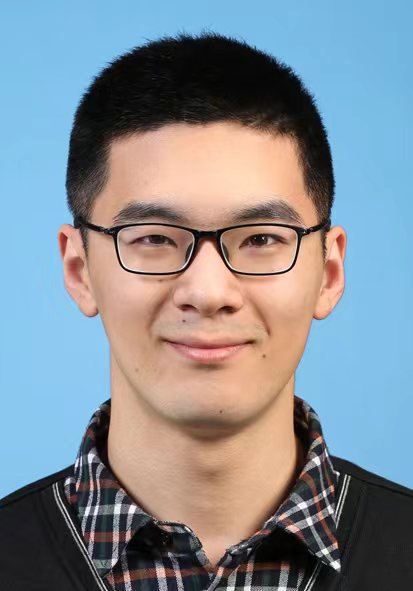}}]{Naiyu Gao} received the B.E. degree in computer electronic and information engineering from Xi’an Jiaotong University (XJTU), Xi’an, China, in 2017. He is currently pursuing the Ph.D. degree with the Institute of Automation, Chinese Academy of Sciences (CASIA), Beijing, China. His research interests include image segmentation, deep learning, pattern recognition, and computer vision.
\end{IEEEbiography}

\begin{IEEEbiography}[{\includegraphics[width=1in,height=1.25in,clip,keepaspectratio]{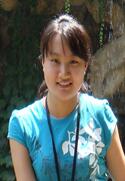}}]{Xiaotang Chen} received the B.E. degree from Xidian University in 2008 and the Ph.D. degree in pattern recognition and intelligent systems from the Institute of Automation, Chinese Academy of Sciences (CASIA), in 2013. In 2013, she joined CASIA as an Assistant Professor. She is currently an Associate Professor at the Center for Research on Intelligent System and Engineering.  Her current researches focus on computer vision and pattern recognition, including object tracking, person re-identification and attribute recognition. She served as the technical program committee member of several conferences.
\end{IEEEbiography}

\begin{IEEEbiography}[{\includegraphics[width=1in,height=1.25in,clip,keepaspectratio]{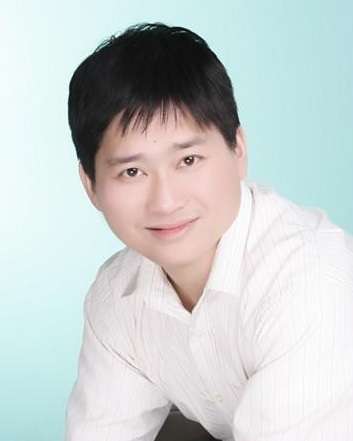}}]{Kaiqi Huang}
 received the B.Sc. and M.Sc. degrees from the Nanjing University of Science Technology, China, and the Ph.D. degree from Southeast University. He is a Full Professor in the Center for Research on Intelligent System and Engineering (CRISE), Institute of Automation, Chinese Academy of Sciences (CASIA). He is also with the University of Chinese Academy of Sciences (UCAS), and the CAS Center for Excellence in Brain Science and Intelligence Technology. He has published over 210 papers in the important international journals and conferences, such as the IEEE TPAMI, T-IP, T-SMCB, TCSVT, Pattern Recognition, CVIU, ICCV, ECCV, CVPR, ICIP, and ICPR. His current researches focus on computer vision,pattern recognition and game theory, including object recognition, video analysis, and visual surveillance. He serves as co-chairs and program committee members over 40 international conferences, such as ICCV, CVPR, ECCV, and the IEEE workshops on visual surveillance. He is an Associate Editor of the IEEE TRANSACTIONS ON SYSTEMS, MAN, AND CYBERNETICS: SYSTEMS and Pattern Recognition.
\end{IEEEbiography}







\end{document}